# BRIDGING THE GLOBAL DIVIDE IN AI REGULATION: A PROPOSAL FOR A CONTEXTUAL, COHERENT, AND COMMENSURABLE FRAMEWORK

Sangchul Park[*]

*Abstract*: As debates on potential societal harm from artificial intelligence (AI) culminate in legislation and international norms, a global divide is emerging in both AI regulatory frameworks and international governance structures. In terms of local regulatory frameworks, the European Union (E.U.), Canada, and Brazil follow a "horizontal" or "lateral" approach that postulates the homogeneity of AI, seeks to identify common causes of harm, and demands uniform human interventions. In contrast, the United States (U.S.), the United Kingdom (U.K.), Israel, and Switzerland (and potentially China) have pursued a "context-specific" or "modular" approach, tailoring regulations to the specific use cases of AI systems. In terms of international governance structures, the United Nations is exploring a centralized AI governance framework to be overseen by a superlative body comparable to the International Atomic Energy Agency. However, the U.K. is spearheading, and the U.S. and several other countries have endorsed, a decentralized governance model, where AI safety institutes in each jurisdiction conduct evaluations of the safety of high-performance general-purpose models pursuant to interoperable standards. This paper argues for a context-specific approach alongside decentralized governance, to effectively address evolving risks in diverse mission-critical domains, while avoiding social costs associated with one-size-fits-all approaches. However, to enhance the systematicity and interoperability of international norms and accelerate global harmonization, this paper proposes an alternative contextual, coherent, and commensurable (3C) framework. To ensure contextuality, the framework (i) bifurcates the AI life cycle into two phases: learning and deployment for specific tasks, instead of defining foundation or general-purpose models; and (ii) categorizes these tasks based on their application and interaction with humans as follows: autonomous, discriminative (allocative, punitive, and cognitive), and generative AI. To ensure coherency, each category is assigned specific regulatory objectives replacing 2010s vintage "AI ethics." To ensure commensurability, the framework promotes the adoption of international standards for measuring and mitigating risks.

---

[*] Assistant Professor, School of Law, Seoul National University (SNU), with joint appointments at the Interdisciplinary Program in Artificial Intelligence and the Department of Mathematical Information Science. I extend my deepest gratitude to the Washington International Law Journal editors for their unwavering support and invaluable advice. This paper was funded by the 2024 Research Fund of the SNU Law Research Institute, donated by the SNU Law Foundation.

# I. INTRODUCTION

The rise and widespread adoption of artificial intelligence (AI) have fueled heated debates over its potential societal harms and legal oversight. Historically, there have been several turning points where scientific progress has extensively influenced the legal system, spurring novel regulatory regimes addressing unprecedented scientific risks. For example, automobiles survived precautionary laws like the United Kingdom (U.K.)'s Locomotive Act of 1865[1] and have integrated into the legal system through innovative tort, product liability, insurance, and safety-regulation regimes. As harms from technology grew, environmental and climate change laws emerged as the quintessence of scientific risk regulation.

In the mid-2010s, when innovative deep learning technologies, such as convolutional neural network (CNN) and generative adversarial network (GAN), stunned the public with their unprecedented image-processing capabilities, debates about "AI ethics" emerged.[2] AI was described as a "black box" that undermined human autonomy and control and made it difficult to identify and address its inherent biases.[3] AI ethics guidelines, focusing on principles like fairness, accountability, and transparency (FAccT), have burgeoned since the late 2010s.[4]

These "soft laws" are transitioning to regulatory laws. The E.U. AI Act has become effective on August 1, 2024, and will be fully applicable two years after its entry into force.[5] Then it will be applied across the E.U., not requiring E.U. member

---

[1] Locomotive Act 1865, 28 & 29 Vict. c. 83 (Gr. Brit.). It required vehicles to be accompanied by at least three persons including a red flag bearer preceding the vehicle and imposed a speed limit of 4 mph (2 mph in towns).

[2] Anna Jobin, Marcello Lenca & Effy Vayena, *The Global Landscape of AI Ethics Guidelines*, 1 NAT. MAC. INTEL. 389 *passim* (2019).

[3] Davide Castelvecchi, *The Black Box of AI*, 538 NATURE 20, 21 (2016); Matthew U. Scherer, *Regulating Artificial Intelligence Systems: Risks, Challenges, Competencies, and Strategies*, 29 HARV. J. L. TECH. 353, 363–64 (2016).

[4] For examples of the AI guidelines, see (1) WHITE HOUSE, BLUEPRINT FOR AN AI BILL OF RIGHTS (2022), https://www.whitehouse.gov/ostp/ai-bill-of-rights; (2) U.K. CENT. DIGIT. & DATA OFF., DATA ETHICS FRAMEWORK (2018) (last amended Sept. 16, 2020), https://www.gov.uk/government/publications/data-ethics-framework; (3) E.C. High-Level Expert Grp. on A.I., Ethics Guidelines for Trustworthy AI (2019) (EU), https://digital-strategy.ec.europa.eu/en/library/ethics-guidelines-trustworthy-ai; (4) OECD, RECOMMENDATION OF THE COUNCIL ON ARTIFICIAL INTELLIGENCE (2019) (last amended May 3, 2024), https://legalinstruments.oecd.org/en/instruments/OECD-LEGAL-0449; and (5) DOCUMENTS OF ACHIEVEMENT, HIROSHIMA AI PROCESS, HIROSHIMA PROCESS INTERNATIONAL GUIDING PRINCIPLES FOR ALL AI ACTORS (Oct. 30, 2023), https://www.soumu.go.jp/hiroshimaaiprocess/en/documents.html.

[5] Council Regulation 2024/1689 of Jun. 13, 2024 laying down harmonized rules on artificial intelligence and amending Regulations (EC) No 300/2008, (EU) No 167/2013, (EU) No 168/2013, (EU) 2018/858, (EU) 2018/1139

countries to devise their own laws. The E.U. AI Act adopts a "horizontal" approach that postulates the homogeneity of AI systems, seeks to identify common causes of harm, and demands uniform human interventions. Canada, Brazil, and South Korea modeled their AI bills after the E.U. AI Act.

In contrast, the U.S., U.K., Israel, and Switzerland (and potentially China) have pursued "context-specific" approaches, which tailor regulations to the specific applications of each AI system. While the concept derives from U.K. policy documents, recent U.S. standards and executive orders are giving it shape. The National Institute of Standards and Technology (NIST)'s Artificial Intelligence Risk Management Framework (AI RMF 1.0) proposes a four-step, context-specific risk management process: (1) "Govern," establishing a risk management system; (2) "Map," recognizing context and identifying risks; (3) "Measure," assessing identified risks; and (4) "Manage," prioritizing and addressing risks.[6] The Biden Administration's E.O. 14110 on the Safe, Secure, and Trustworthy Development and Use of AI identifies risks, especially those pertaining to national security and public safety, in various domains and designates specific roles for agencies.[7]

The fragmentation of AI regulatory regimes should be addressed to ensure the cross-border interoperability of AI and lay a solid foundation for international governance. Preparations for an international governance regime are already in progress. The United Nations (U.N.) is exploring a centralized governance framework to be overseen by a superlative body comparable to the International Atomic Energy Agency (IAEA). The U.N. High-Level Advisory Body on AI issued an interim report on international governance options[8] on December 31, 2023, in which it presented five guiding principles[9] and seven institutional functions.[10] The

---

and (EU) 2019/2144 and Directives 2014/90/EU, (EU) 2016/797 and (EU) 2020/1828 (Artificial Intelligence Act), 2016 O.J. (L 119) 1–88 (EU), 2024 O.J. (L 1689) (EU).

[6] NAT'L INST. STANDARDS & TECH. (NIST), NIST AI 100-1, ARTIFICIAL INTELLIGENCE RISK MANAGEMENT FRAMEWORK (AI RMF 1.0) (2023).

[7] Exec. Order No. 14,110, 88 Fed. Reg. 75191 (Nov. 1, 2023).

[8] U.N. AI ADVISORY BODY, INTERIM REP.: GOVERNING AI FOR HUMANITY (2023).

[9] Guiding principles include (1) inclusiveness, (2) public interest, (3) data governance and data commons, (4) multi-stakeholder collaboration, and (5) alignment with international agreements. *Id.* at 13–15.

[10] Institutional functions include (1) horizontal scanning, building scientific consensus, (2) interoperability and alignment with norms, (3) mediating standards, safety, and risk management frameworks, (4) facilitation of development and use—liability regimes, cross-border model training and testing, (5) international collaboration on data, compute, and talent to solve sustainable development goals, (6) reporting and peer review, and (7) norm elaboration, compliance, and accountability. *Id.* at 15–18.

U.N. plans to finalize the report in the summer of 2024, and issue another report on how to organize the superlative body. On the other hand, the U.K. is spearheading, and the U.S. and several other countries have endorsed, a decentralized governance model, where AI Safety Institutes (AISIs) in each jurisdiction conduct evaluations of the safety of high-performance general-purpose models (termed "frontier AI") pursuant to interoperable standards.[11] Its outline titled the Bletchley Declaration was endorsed at the 1st AI Safety Summit in November 2023,[12] and was reaffirmed at the AI Seoul Summit in May 2024. The U.K. launched the world's first AISI in November 2023, tasking it with (i) developing and conducting evaluations on advanced AI systems; (ii) driving foundational AI safety research; and (iii) facilitating information exchange.[13] The U.S. and Japan have established, and South Korea and Canada are establishing, their own AISIs.[14]

However, there is still no consensus regarding how to materialize a globally interoperable regulatory framework. While the E.U.'s approach aligns with its commitment to human rights, its AI Act lacks the necessary level of proportionality, granularity, and foreseeability to effectively address evolving risks in diverse mission-critical domains. Pro-innovation jurisdictions basing their regulations on cost-benefit analyses cannot justify requiring "high-risk" AI to comply with all safety, fairness, accountability, accuracy, robustness, and privacy regulations when AIs involving safety risks, biases, infringements, quality-of-service (QoS) harms, and privacy problems can be legislated separately. The context-specific approach holds greater promise in this respect, but it requires further development of details, coherent regulatory objectives, and commensurable metrics to be considered an international norm.

---

[11] The basic concept of the safety of frontier AI was coined by Anthropic (ANTHROPIC, CORE VIEWS ON AI SAFETY: WHEN, WHY, WHAT, AND HOW (Mar. 8, 2023), https://www.anthropic.com/news/core-views-on-ai-safety); endorsed by Microsoft, Anthropic, Google, and OpenAI (MICROSOFT, MICROSOFT, ANTHROPIC, GOOGLE, AND OPENAI LAUNCH FRONTIER MODEL FORUM (Jul. 26, 2023), https://blogs.microsoft.com/on-the-issues/2023/07/26/anthropic-google-microsoft-openailaunch-frontier-model-forum); and emerged as a key agenda at the AI Safety Summit spearheaded by the U.K. (U.K. DEPT. SCI., INNOVATION &TECH. (DSIT), CAPABILITIES AND RISKS FROM FRONTIER AI (Nov. 1–2, 2023), https://www.gov.uk/government/publications/frontier-ai-capabilities-and-risks-discussion-paper).

[12] THE BLETCHLEY DECLARATION BY COUNTRIES ATTENDING THE AI SAFETY SUMMIT (UK) (Nov. 1–2, 2023), https://www.gov.uk/government/publications/ai-safety-summit-2023-the-bletchley-declaration/the-bletchley-declaration-by-countries-attending-the-ai-safety-summit-1-2-november-2023.

[13] U.K. DSIT, INTRODUCING THE AI SAFETY INSTITUTE (Nov. 2023), https://www.gov.uk/government/publications/ai-safety-institute-overview/introducing-the-ai-safety-institute.

[14] NIST, U.S. COMMERCE SECRETARY GINA RAIMONDO ANNOUNCES KEY EXECUTIVE LEADERSHIP AT U.S. AI SAFETY INSTITUTE (Feb. 7, 2024), https://www.nist.gov/news-events/news/2024/02/us-commerce-secretary-gina-raimondo-announces-key-executive-leadership-us; INFO. TECH. PROMOTION AGENCY, JAPAN, PRESS RELEASE – ESTABLISHING THE AI SAFETY INSTITUTE (Feb. 14, 2024), https://www.ipa.go.jp/pressrelease/2023/press20240214.html; MINISTRY SCI. & ICT, KOREA (MSIT), ANNOUNCING THE MSIT'S KEY POLICIES IN 2024 (Feb. 13, 2024), https://www.msit.go.kr/bbs/view.do?mId=113&mPid=238&bbsSeqNo=94&nttSeqNo=3184072; DEP'T FIN. CAN., REMARKS BY THE DEPUTY PRIME MINISTER ON SECURING CANADA'S AI ADVANTAGE (Apr. 7, 2024), https://www.canada.ca/en/department-finance/news/2024/04/remarks-by-the-deputy-prime-minister-on-securing-canadas-ai-advantage.html.

As a tool for bridging the divide, this paper proposes a contextual, coherent, and commensurable (3C) framework. Grounded in a context-specific approach, the 3C framework aims to enhance the regulatory framework's coherence by classifying AIs according to their model types and phases. Based on the classification, the 3C framework takes the following three steps.

First, to ensure contextuality, the framework bifurcates the AI life cycle into two phases: (1) learning of models and (2) deployment of the models for specific tasks, instead of specifically defining foundation or general-purpose models. It further categorizes these tasks based on their usage and interaction with humans: autonomous, discriminative (allocative, punitive, and cognitive), and generative AI (See Figure 1).

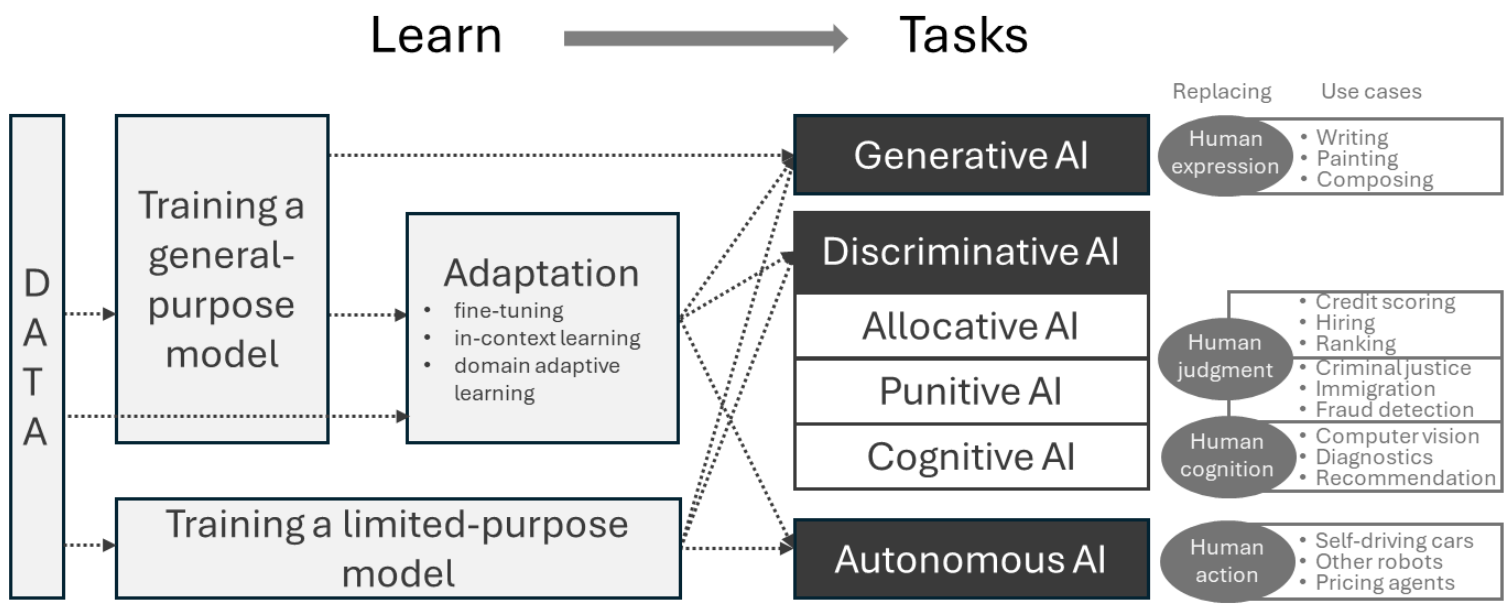

*Figure 1 Categorizing AI phases and tasks (detailed in Section IV)*

Second, to ensure *coherency*, each "task" category is assigned regulatory objectives that replace 2010s vintage "AI ethics": (1) safety for autonomous AI; (2) fairness and explainability for allocative AI; (3) accuracy and explainability for punitive AI; (4) accuracy, robustness, and privacy for cognitive AI; and (5) the mitigation of infringement and misuse for generative AI. The framework further addresses the learning process, focusing on its role as the "harvest of knowledge" and optimizing liability related to data mining, copyright, and personal data.

Third, to ensure *commensurability* regarding each regulatory objective, the framework supports adopting international industry standards for measuring and mitigating evolving risks. This scheme aligns well with the decentralized governance structure, where each AISI conducts model evaluations based on international standards and mutually recognizes the outcomes of the evaluations.

With the above as a framework, this paper is structured as follows. Section II provides an overview of the existing AI regulatory frameworks in various jurisdictions, focusing on the global divide between the horizontal and context-specific approaches. Section III examines the suitability of each approach as an international norm. Section IV analyzes distinct societal harms arising from different phases and types of AIs, proposes tailored regulatory objectives for each type, and calls for the adoption of international industry standards that convert the principles into quantifiable metrics. Section V concludes.

## II. GLOBAL DIVIDE IN AI REGULATION: HORIZONTAL V. CONTEXT-SPECIFIC

The horizontal and context-specific approaches characterize the global divide concerning AI regulation. The terms "horizontal" and "context-specific" are credited to the explanatory memorandum within the E.U. AI Act and the U.K.'s post-Brexit AI policy,[15] respectively. Israel refers to them as "lateral" versus "modular." They can alternatively be called "bundled" versus "unbundled,"[16] "omnibus" versus "sectoral,"[17] or "horizontal layered" versus "vertical silo."[18]

The horizontal approach assumes AI systems are homogeneous and induce common societal harm. The E.C.'s impact assessment for the legislation of the AI Act identifies (1) opacity, (2) complexity, (3) autonomous behavior, (4) continuous adaptation and unpredictability, and (5) data dependency as the "problem drivers" justifying a uniform package of regulations for high-risk AI systems.[19] To cope with these shared causes of harm, it calls for omnibus AI laws that apply a full, uniform package of rules regarding fairness, transparency, accountability, human oversight, accuracy, robustness, and security. Since the E.U. prepared its AI Act as a horizontal framework, Canada, Brazil, and South Korea have begun preparing laws modeled after this framework.

---

[15] U.K. DSIT & OFFICE FOR A.I. (OAI), A PRO-INNOVATION APPROACH TO AI REGULATION 25 (2023), https://www.gov.uk/government/publications/ai-regulation-a-pro-innovation-approach/white-paper (stating "[o]ur framework is context-specific. We will not assign rules or risk levels to entire sectors or technologies. Instead, we will regulate based on the outcomes AI is likely to generate in particular applications.").
[16] Jennifer Nou & Edward H. Stiglitz, *Regulatory Bundling*, 128 YALE L. J. 1174 *passim* (2019).
[17] Paul M. Schwartz, *Preemption and Privacy*, 118 YALE L. J. 806, 908–15 (2009) (discussing privacy laws).
[18] Richard S. Whitt, *A Horizontal Leap Forward: Formulating a New Communications Public Policy Framework Based on the Network Layers Model*, 56 FED. COMM. L. J. 587, 590–91 (2004) (discussing telecommunications regulations).

Conversely, context-specific approaches regulate AI systems according to their use cases and their effects on people within specific contexts.[20] Countries leading AI research and development, including the U.S., U.K., and Israel (and potentially China), tend to adopt the context-specific approach.[21] This approach believes AI-induced societal harms mirror existing problems in various domains where AI has replaced or complemented traditional methodologies. Even if a novel societal harm arises from a particular context where AI is used, this approach prefers to address it with tailored regulations. The approach thus favors adapting the legal system to societal harms in different contexts rather than enacting an umbrella AI regulation.

### 1. Horizontal Approaches

### A. The E.U. AI Act

Before the E.U. AI Act was proposed, the General Data Protection Regulation (GDPR) granted data subjects the right not to be subject to a decision based solely on automated processing.[22] This provision was transplanted into a host of automated decision-making (ADM) laws in other jurisdictions, including the U.S.—at the state level, as noted below. As U.S. big techs dominate the European digital platform market, the E.U. enacted *ex ante* platform regulations that expanded disclosure obligations for ranking and recommender systems.[23]

---

[19] E.C., *Commission Staff Working Document Accompanying the Proposal for a Regulation of the European Parliament and of the Council Laying Down Harmonized Rules on Artificial Intelligence (Artificial Intelligence Act) and Amending Certain Union Legislative Acts*, at 28–30, COM (2021) 84 final (April 21, 2021).

[20] U.K. DEP'T. FOR BUS., ENERGY & INDUS. STRATEGY (BEIS) ET AL., ESTABLISHING A PRO-INNOVATION APPROACH TO REGULATING AI (2022), https://www.gov.uk/government/publications/establishing-a-pro-innovation-approach-to-regulating-ai/establishing-a-pro-innovation-approach-to-regulating-ai-policy-statement.

[21] Top ten countries in private AI investment during 2013 to 2023 are the U.S. ($335.24B), China ($103.65B), the U.K. ($22.25B), Israel ($12.83B), Canada ($10.56B), Germany ($10.35B), India ($9.85B), France ($8.31B), South Korea ($7.25B), and Singapore ($6.25B). A.I. INDEX & STAN. UNIV. HUMAN-CENTERED A.I., ARTIFICIAL INTELLIGENCE INDEX REPORT 248 (2024).

[22] Council Regulation 2016/679 of Apr. 27, 2016 on the Protection of Natural Persons with regard to the Processing of Personal Data and on the Free Movement of Such Data, and Repealing Directive 95/46/EC (General Data Protection Regulation), art. 22, 2016 O.J. (L 119) 1–88 (EU).

Realizing these regulations' limitations, the E.U. developed an omnibus regulatory framework. The AI Act categorizes uses of AI based on risk level and applies regulations of varying stringency. It bans "prohibited" AI practices,[24] subjects "high-risk" AI systems[25] to stringent *ex ante* regulations,[26] and imposes transparency obligations over intermediate-risk AIs.[27] The E.U. assumes "high-risk" AIs share common "problem drivers," necessitating a full, uniform package of rules on fairness, transparency, accountability, human oversight, accuracy, robustness, and security. For instance, while existing laws expect elevators to be safe and recruiting or credit scoring processes to be fair and transparent, the AI Act demands both fair and transparent self-driving elevators and safe hiring or credit scoring AIs merely because AI is used. After the advent of ChatGPT, new regulations over general-purpose AI (GPAI) models were added.[28]

---

[23] Under the Platform-to-Business Act, online intermediation service providers and online search engines must disclose the main parameters for ranking determinations and their relative importance. Regulation 2019/1150 of Jun. 20, 2019 on Promoting Fairness and Transparency for Business Users of Online Intermediation Services, art. 5, 2019 O.J. (L 186) 57–79 (EU). Under the Digital Services Act, online platforms must disclose the main parameters used in their recommender systems and options for service recipients to modify or influence those parameters. Regulation 2022/2065 of Oct. 19, 2022 on a Single Market for Digital Services and Amending Directive 2000/31/EC (Digital Services Act), art. 27, O.J. (L 277) 1–102 (EU).

[24] The prohibited AI practices include (1) deploying subliminal techniques; (2) exploiting the vulnerabilities of minors, disabled persons, or those under a specific social or economic situation; (3) social scoring; (4) assessing the risk of criminal offense; (5) building facial recognition databases through untargeted scraping; (6) inferring emotion in the workplace or school; (7) biometric categorization for inferring sensitive data; (8) real-time remote biometric identification in public spaces for law enforcement (art. 5(1)(a) to (h)).

[25] The high-risk AI systems include products or safety components covered by the Union harmonization legislation, including vehicles and other transportation means, elevators, machinery, medical devices, toys, etc. (art. 6(1), Annex I). They further include (1) biometrics; (2) critical infrastructure; (3) education and vocational training; (4) employment; (5) essential services and benefits (evaluating eligibility for public assistance, credit scoring, insurance underwriting, and evaluating emergency calls); (6) law enforcement; (7) migration, asylum, and border control management; and (8) administration of justice and democratic processes (art. 6(2), Annex III). However, an AI system is not deemed high-risk if it does not pose a "significant risk of harm to the health, safety or fundamental rights of natural persons" (art. 6(3)).

[26] High-risk AI systems must undergo conformity assessment and comply with AI system requirements, including risk management systems; data and data governance; technical documentation; record-keeping; transparency and providing information to deployers; human oversight; and accuracy, robustness, and cybersecurity (arts. 8 to 15). Providers, importers/distributors, and deployers of the high-risk AI systems are subject to different obligations (arts. 16 to 27). For example, providers should comply with the AI system requirements; disclose provider information; have in place quality management systems; prepare and keep technical documentation; keep logs; undergo conformity assessments; implement declarations of conformity; affix CE marking (conformity certification marking); register with the E.U. database; take corrective actions and provide information; demonstrate conformity at authorities' request; ensure accessibility; appoint authorized representatives (for an offshore operator); implement post-market monitoring; and report serious incidents (arts. 16, 22, 72, and 73). As an exception, those listed in Section B of Annex I are exempted from most provisions of the EU AI Act (art. 2(2)), and those listed in Section A of Annex I have a choice of integrating the necessary testing and reporting processes, information, and documentation under the EU AI Act into those under the Union harmonization legislation (art. 8(2)).

Each regulatory provision is stringent, necessitating substantial compliance costs. For example, the technical documentation, which high-risk AI system providers are required to prepare, keep, and submit to conformity assessment bodies, includes extensive volumes of information.[29] The deployers of biometric identification must assign at least two natural persons to verify and confirm the system's actions or decisions.[30] Under the AI Act, GPAI model providers must disclose a "sufficiently detailed summary about the content used for training [the GPAI]."[31] Given the size of corpora used for contemporary large language models, it is impossible to individually list copyrighted materials. For example, Common Crawl, a corpus used to train various language models, such as GPT-3, T5, and Gopher, includes 454 TiB

---

[27] People should be informed when interacting with AI systems involving human interaction, emotion recognition or biometric categorization, and deepfakes (art. 50(1), (3), and (4)). The providers of generative AI must ensure outputs be marked in a machine-readable format, and detectable as artificially generated or manipulated (art. 50(2)).

[28] Providers of GPAI models must keep technical documentation; provide information and documentation to downstream providers for compliance; implement a copyright policy; and disclose a summary of the content used for training (art. 53). Providers of GPAI models with systemic risks must additionally conduct model evaluations; risk assessments and risk mitigation measures; adequate cybersecurity protection; and the reporting of serious incidents (art. 55).

[29] The technical documentation requirement includes: (1) a general description of the AI system; (2) development methods and steps; (3) design specifications; (4) system architecture; (5) data (including training methodologies, training data sets, provenance, scope and characteristics, data sources, labeling, and data cleaning); (6) human oversight; (7) pre-determined changes; (8) validation and testing procedures (including the training data used and its characteristics, metrics measuring accuracy, robustness, and compliance with other requirements, and test logs and reports); (9) cybersecurity measures, (10) monitoring, functioning, and AI control information; (11) the appropriateness of performance metrics; (12) risk management systems; (13) relevant provider-made changes; (14) a list of applicable harmonized standards; (15) a copy of the declaration of conformity; and (16) system description for post-market monitoring. Art. 11, Annex IV.

[30] Art. 14(5).

[31] Art. 53(1)(d).

(about 500 trillion bytes) of data sourced from 3.35 billion webpages, from November to December 2023.[32]

## B. Canada: Artificial Intelligence and Data Act (AIDA)

Canada is embracing AI within its Digital Charter by adopting a more lenient horizontal framework as compared to the EU AI Act.[33] Canada introduced the three-bill Digital Charter Implementation Act 2022 (Bill C-27) in 2022.[34] These bills are being considered in the Standing Committee on Industry and Technology within the House of Commons. One bill is AIDA, which focuses on the accountability of persons responsible for a high-impact system—that is, those who design, develop, or make the system available for use or manage its operation. The government will issue regulations defining the criteria for assessing whether an AI system is a high-impact system.[35] Those responsible for a high-impact system must establish measures to identify, assess, and mitigate the risks of harm or biased output caused by the system's use; must monitor mitigation measure compliance and effectiveness; and should keep records, publish a description of the system, and notify material harms to the government.[36]

## C. Brazil: AI Bill (*Projeto de Lei nº 2338, de 2023*)

Brazil has imposed stringent regulations over U.S. Big Techs and is now preparing the most stringent version of the E.U. AI Act. The Chamber of Deputies' (*Câmara dos Deputados*) lower house approved the AI Bill (*Projeto de Lei nº 21, de 2020*) in 2021. However, the Senate (*Senado Federal*) found the bill too lenient and organized a legal expert commission (CJSubIA) to prepare a December 2022 report and draft a more stringent version. CJSubIA submitted its revised AI bill (*Projeto de Lei nº 2338, de 2023*) to the Senate in May 2023.[37]

---

[32] Thom Vaughan, *November/December 2023 Crawl Archive Now Available* (Dec. 15, 2023), https://www.commoncrawl.org/blog/november-december-2023-crawl-archive-now-available.

[33] The Charter in Canada refers to the rights provisions that constitute the first 35 sections of the Constitution Act of 1982. Building upon this concept, Canada introduced the Digital Charter in 2019, which was developed based on the input of citizens. The Digital Charter consists of ten principles, including: (1) universal access, (2) safety and security, (3) control and consent, (4) transparency, portability, and interoperability, (5) open and modern digital government, (6) level playing field, (7) data and digital for good, (8) strong democracy, (9) exclusion of hate and violent extremism, and (10) strong enforcement and real accountability. CANADA'S DIGITAL CHARTER IN ACTION: A PLAN BY CANADIANS, FOR CANADIANS (Oct. 23, 2019), https://www.ic.gc.ca/eic/site/062.nsf/eng/h_00109.html.

[34] House of Commons of Canada Bill C-27, An Act to Enact the Consumer Privacy Protection Act, the Personal Information and Data Protection Tribunal Act and the Artificial Intelligence and Data Act and to make consequential and related amendments to other Acts, June 16, 2022, https://www.parl.ca/legisinfo/en/bill/44-1/c-27; THE ARTIFICIAL INTELLIGENCE AND DATA ACT (AIDA) - COMPANION DOCUMENT (Mar. 13, 2023), https://ised-isde.canada.ca/site/innovation-better-canada/en/artificial-intelligence-and-data-act-aida-companion-document.

[35] *Id*.

[36] *Id*.

Unlike the E.U., the bill regulates all AI systems in the following areas: risk assessment; individual rights; governance; shifting the burden of proof; codes of conduct; and security breach notifications.[38] However, it bans excessive risk AI systems like the E.U. and strengthens regulations for high-risk AI systems, with regard to stricter governance measures (including technical measures for explainability and designation of a team to ensure diverse viewpoints); an algorithmic impact assessment; public databases; and strict liability.[39]

### D. South Korea: Proposal for the Act on the Development of and Establishing a Basis of Trust in AI

South Korea's AI bill, which consolidated seven preceding bills in February 2023, required business operators related to "high-risk domain AIs" to inform users that a product or service is operated based on the high-risk domain AI and to take measures to ensure the trustworthiness and safety of AI. "High-risk domain AIs" include: (1) judging or scoring AIs significantly impacting individual rights and obligations; (2) AIs used by public agencies for decision-making that affects citizens; (3) AIs for analysis and use of biometric data for investigations or arrests; and (4) AI used for energy supply, drinking water supply, public health, medical devices, nuclear facilities, and traffic.

However, this legislative initiative has encountered internal objections. Proponents of innovation criticize the bill for replicating the E.U.'s approach without recognizing the Korean tech industry's potential to survive through innovation rather than resorting to legal barriers or regulatory ecosystems.[40] Meanwhile, human rights advocates argue that the bill should adhere more closely to the E.U. AI Act.[41] The

---

[37] PROJETO DE LEI N° 2338, DE 2023 (Braz.), https://www25.senado.leg.br/web/atividade/materias/-/materia/157233.
[38] *Id.*, Arts. 7–13, 19, 27, 30 & 31.
[39] *Id.*, Arts. 14–16, 20–27 & 43.
[40] Sangchul Park, Opinion, *AI Regulation: The EU Model is Not a Right Answer to It*, JOONGANG DAILY, Dec. 14, 2023 (S. Kor.), https://www.joongang.co.kr/article/25214651.
[41] The National Human Rights Commission (NHRC) requested, in July 2023, that the AI bill align more closely with the EU AI Act, issuing an opinion stating that the bill should undergo further revisions to strengthen the rights and remedies of users and information subjects, expand the scope of high-risk domain AIs, implement *ex ante* regulations and suspension orders for high-risk domain AIs, introduce human rights impact assessments, and establish an

bill expired on May 29, 2024, when a legislation session ended, but was resubmitted at the current session. Efforts are underway to explore alternatives.[42]

## 2. Context-Specific Approaches

### A. United Kingdom: "Pro-Innovation" and "Context-Specific" Approach

The U.K. has rejected the E.U. approach and instead proposed a pro-innovation, context-specific framework. The Department for Digital, Culture, Media, and Sport (DCMS) submitted to Parliament a white paper titled "Establishing a Pro-Innovation Approach to Regulating AI" in 2022.[43] That report emphasized that, compared to the E.U.'s rigid framework based on product safety regulations, Brexit facilitated the U.K.'s creation of a nimbler regulatory system.[44] The report also criticized the E.U. approach as unsuitable for the U.K. because it ignores the application of AI and its regulatory implications, and hinders innovation due to a lack of granularity.[45] As an alternative, the report proposed four characteristics of the pro-innovation approach: (1) context-specific, (2) pro-innovation and risk-based, (3) coherent, (4) and proportionate and adaptable.[46] The Department for Science, Innovation and Technology (DSIT) submitted to Parliament a white paper titled, "Pro-Innovation Approach to AI Regulation" in 2023.[47] It aligns with the perspective presented in the DCMS's white paper and outlines several policy initiatives.[48] The policy initiatives include (1) monitoring and evaluating the regulatory framework's overall effectiveness; (2) assessing and monitoring AI-induced risks across the economy; (3) conducting horizon scanning and gap analysis; (4) supporting testbeds and sandbox

---

independent regulatory authority. NHRC, AI BILL – THE GO-FIRST-REGULATE-LATER PRINCIPLE SHOULD BE ERASED AND THE HUMAN RIGHT IMPACT ASSESSMENT SHOULD BE INTRODUCED (Aug. 24, 2023), https://www.humanrights.go.kr/base/board/read?boardManagementNo=24&boardNo=7609439.

[42] The Personal Information Protection Committee (PIPC) set forth its plan to establish a flexible and adaptable regulatory framework. PIPC, POLICY DIRECTION FOR SAFE USE OF PERSONAL DATA IN THE AI ERA (Aug. 2023), https://www.pipc.go.kr/np/cop/bbs/selectBoardArticle.do?bbsId=BS074&mCode=C020010000&nttId=9083.

[43] *See* U.K. BEIS ET AL., *supra* note 20.

[44] *Id.*

[45] *Id.*

[46] *Id.*

[47] *See* U.K. DSIT & OAI, *supra* note 15.

[48] *Id.*

initiatives; (5) educating and raising awareness; and (6) promoting interoperability with international regulatory frameworks.[49]

Apart from these department-specific policy initiatives, Downing Street spearheaded the AI Safety Summit to promote an idea of controlling the safety of "frontier AI."[50] As noted, this initiative has the potential for evolving into decentralized international control regime implemented by AISIs in each jurisdiction.

### B. United States

#### (1) AI RMF 1.0 and E.O. 14110

While the context-specific approach is credited to the U.K., the NIST's AI RMF 1.0 and the Biden Administration's E.O. 14110 have shaped this approach. AI RMF 1.0 delineates a four-step process for risk management: Govern-Map-Measure-Manage. The "Map" function is a good example of how to shape the context-specific approach in a real AI life cycle. It has five key steps: (1) establishing and comprehending the context; (2) categorizing the AI system; (3) understanding AI capabilities, intended usage, objectives, and anticipated benefits and costs; (4) mapping risks and benefits; and (5) characterizing effects on individuals and society.[51]

Under E.O. 14110, more than 50 federal entities must undertake over 100 actions to implement the guideline provided across 8 overarching policy areas: (1) promoting safety and security; (2) fostering innovation and competition; (3) supporting workers impacted by AI adoption; (4) addressing AI bias and civil rights considerations; (5) protecting consumer interests; (6) privacy safeguards; (7) coordinating federal AI use; and (8) international engagement and standardization efforts.[52] Among them, Section 4 provides 27 safety and security requirements governing over 30 federal entities.[53] Additionally, this section addresses the control of "dual-use foundation models," which are general-purpose models with a model size of at least 10 billion parameters and which present significant risks to security, national economic security, national public health or safety, or any combination of those matters.[54] The Department of Commerce (DOC) is tasked with (1) mandating developers of dual-use foundation models to regularly report model training, parameters, and safety measures, including red-teaming tests, to the federal government;[55] (2) discussing and reporting strategies for handling widely available dual-use foundation models, which can impede export control efforts;[56] and (3) establishing the NIST guidelines for AI red-teaming tests.[57] Additionally, holders of high-performance AI models (those exceeding $10^{26}$ integer or floating-point operations, or $10^{23}$ for biological sequence data) and high-performance computing clusters (with data center networking surpassing 100 Gbit/s and $10^{26}$ integer or floating-point operations per second) must report to the federal government.[58] In addition, large AI models (exceeding $10^{26}$ integer or floating-point operations) must track foreign transactions and assist in investigating foreign cyber attackers.[59] This suggests that the U.S.'s national security policy towards AI is transitioning from human control over the Lethal Autonomous Weapons Systems (LAWS) to export control over dual-use foundation models and high-performance computing infrastructure.[60] Section 4 also encompasses comprehensive mandates aimed at controlling the use of AI in cyberattacks and chemical, biological, radiological, and nuclear (CBRN) threats[61] and ensuring digital content authentication and synthetic content detection for generative AI.[62]

---

[49] *Id.* at § 14.
[50] *See* sources cited *supra* note 11.
[51] *See* NIST, *supra* note 6, at 24–28.
[52] 88 Fed. Reg. 75191, §§ 4–11.
[53] *Id.* at § 4.

### (2) Federal Legislation

---

[54] *Id.* at § 3(k).
[55] *Id.* at § 4.2(a)(i).
[56] *Id.* at § 4.6.
[57] *Id.* at § 4.1(a)(ii).
[58] *Id.* at § 4.2(b).
[59] *Id.* at § 4.2(c).
[60] The current international export control system is based on the U.S.-led four multilateral regimes: the Wassenaar Arrangement (WA), the Nuclear Suppliers Group (NSG), the Australia Group (AG), and the Missile Technology Control Regime (MTCR). Given that Russia is a member of three agreements excluding the AG, the Russia-Ukraine War underscored the need for restructuring the current regime, and AI, particularly foundation models, is going to be seriously incorporated into the alternative export control regime.
[61] 88 Fed. Reg. 75191, § 4.4, § 4.7(a). *See also* DEPT. OF HOMELAND SECURITY, FACT SHEET: DHS ADVANCES EFFORTS TO REDUCE THE RISKS AT THE INTERSECTION OF ARTIFICIAL INTELLIGENCE AND CHEMICAL, BIOLOGICAL, RADIOLOGICAL, AND NUCLEAR (CBRN) THREATS (Apr. 29, 2024), https://www.dhs.gov/sites/default/files/2024-04/24_0429_cwmd-dhs-fact-sheet-ai-cbrn.pdf.
[62] *Id.* at § 4.5. See also NIST, NIST AI 100-4, REDUCING RISKS POSED BY SYNTHETIC CONTENT, Draft for Public Comment (2024), https://airc.nist.gov/docs/NIST.AI.100-4.SyntheticContent.ipd.pdf.

The FTC continues to enforce AI-related issues, following Section 5 of the FTC Act, the Fair Credit Reporting Act (FCRA), and the Equal Credit Opportunity Act (ECOA).[63] Bills for the Algorithmic Accountability Act (AAA) plan to grant the FTC comprehensive administrative and enforcement authority regarding AI regulation.[64] The AAA was initially introduced in 2019 and was reintroduced in 2022 and 2023. Although lenient compared to the E.U. AI Act, the AAA shares a horizontal approach coupled with a unitary definition of "critical" AIs.[65] In 2023, the FTC, the Civil Rights Division of the DOJ, the Consumer Financial Protection Bureau (CFPB), and the EEOC released a joint statement outlining their commitment to enforce laws and regulations that ensure fairness, equity, and justice for automated systems.[66]

Following the Senate Judiciary Subcommittee's "Oversight of A.I." hearings in 2023,[67] Senators Richard Blumenthal and Josh Hawley proposed a framework for AI regulation in the U.S. that contained five objectives: (1) establish a licensing regime to be administered by an independent oversight body; (2) ensure legal accountability for AI-related harms; (3) defend national security and international competitiveness; (4) promote transparency; and (5) protect consumers and kids.[68]

---

[63] The FTC has the authority to address AI-related fairness or transparency problems, in accordance with (1) Section 5 of the FTC Act, which prohibits unfair or deceptive acts or practices; (2) the ECOA, which prohibits credit discrimination based on race, color, religion, national origin, sex, marital status, age, or public assistance; and (3) the FCRA, which mandates transparency and accuracy in consumer credit reports. Andrew Smith, FTC Bureau of Consumer Protection, *Using Artificial Intelligence and Algorithms*, FTC BUSINESS BLOG (Apr. 8, 2020), https://www.ftc.gov/business-guidance/blog/2020/04/using-artificial-intelligence-and-algorithms.

[64] S.2892, 118TH CONG. (2023–24), H.R.5628, 118TH CONG. (2023–24).

[65] The AAA requires the FTC to promulgate regulations requiring each covered entity to (1) perform an impact assessment of any "augmented critical decision process" (ACDP)—meaning a process employing an automated decision system to make a "critical decision" —and deployed automated decision systems for ACDP; (2) attempt to eliminate or mitigate the negative effects of ACDPs; and (3) identify capabilities and other resources for improving automated decision systems, ACDPs, or their impact assessment regarding the accuracy, robustness, fairness, transparency, security, safety, and other values. *Id.*

[66] FTC ET AL., JOINT STATEMENT ON ENFORCEMENT EFFORTS AGAINST DISCRIMINATION AND BIAS IN AUTOMATED SYSTEMS (Apr. 25, 2023), https://www.ftc.gov/news-events/news/press-releases/2023/04/ftc-chair-khan-officials-doj-cfpb-eeoc-release-joint-statement-ai.

[67] U.S. Senate Committee on the Judiciary, Subcommittee on Privacy, Technology, and the Law, *Oversight of A.I.: Rules for Artificial Intelligence* (May 16, 2023), *Oversight of A.I.: Principles for Regulation* (July 25, 2023) & *Oversight of A.I.: Legislating on Artificial Intelligence* (Sept. 12, 2023), https://www.judiciary.senate.gov/committee-activity/hearings.

[68] Richard Blumenthal, *Blumenthal & Hawley Announce Bipartisan Framework on Artificial Intelligence Legislation* (Sep. 8, 2023), https://www.blumenthal.senate.gov/newsroom/press/release/blumenthal-and-hawley-announce-bipartisan-framework-on-artificial-intelligence-legislation.

### (3) State Laws

Attorneys general can enforce unfair and deceptive trade practice laws over AI-related practices.[69] Several regulatory statutes have also been enacted, including ADM regulations modeled after the E.U. GDPR. For example, the California Privacy Rights Act of 2020 (CPRA) grants customers subject to ADM the right to opt out, as well as access to information about the "logic involved in those decision-making processes, along with a "description of the likely outcome of the process."[70] Other examples of the ADM regulations include the Colorado Privacy Act,[71] the Virginia Consumer Data Protection Act,[72] and the Connecticut Data Protection Act.[73]

Other sector-specific state statutes include (1) regulations over hiring AI,[74] (2) regulations regarding using chatbots for public elections or marketing,[75] (3) regulations regarding using deepfakes for public election,[76] and (4) facial recognition and biometric data regulations.[77]

### C. Israel: "Modular" and "Sector-Specific" Approach

Israel published a draft of the "principles of the policy for the responsible development of the field of AI" in 2022, marking Israel's first AI regulatory policy initiative.[78] In the report, Israel stated its intent to continue leading AI development by excluding AI-specific "lateral" regulatory frameworks. Instead, Israel intended to utilize soft and advanced regulatory tools in a "modular" format.[79]

---

[69] *See, e.g.,* State v. Clearview AI, Inc., Trial Ord. No. 226-3-20 (Vt. Sup. Ct. Mar. 10, 2020).
[70] CAL. CIV. CODE § 1798.185(a)(16) (2020).
[71] COLO. REV. STAT. §§ 6-1-1306(1)(a)(I)(C), 6-1-1309(1)(2)(a) (2021).
[72] VA. CODE ANN. § 59.1–577(A)(5) (2021), 59.1–580(A)(3) (2021).
[73] 2022 CONN. ACTS 22-15. (Reg. Sess.)
[74] *See, e.g.,* Illinois Artificial Intelligence Video Interview Act, 820 ILL. COMP. STAT. 42/1 *et seq.* (2020)., *and* Maryland Facial Recognition Services Law, MD. CODE ANN, LAB. & EMPL. § 3-717 (2020).
[75] California's 2019 Bolstering Online Transparency (B.O.T.) Act, CAL. BUS. & PROF. CODE § 17941 (2018).
[76] *See, e.g.,* TEX. ELEC. CODE ANN. § 255.004(d) (2021); CAL. ELEC. CODE § 20010(a) (2020) (amended 2022); MINN. STAT. §§ 609.771, 617.262 (2023); WASH. REV. CODE. § 42.62 (2023).
[77] Biometric Information Privacy Act, 740 ILL. COMP. STAT. 14/5 (2008); Capture or Use of Biometric Identifier Act, TEX. BUS. & COM. CODE § 503.001 (2022); Facial Recognition Technology Law, VA. CODE § 15.2-1723.2 (2021).
[78] MINISTRY INNOVATION, SCI. &TECH. (MIST), FOR THE FIRST TIME IN ISRAEL: THE PRINCIPLES OF THE POLICY FOR THE RESPONSIBLE DEVELOPMENT OF THE FIELD OF ARTIFICIAL INTELLIGENCE WERE PUBLISHED FOR PUBLIC COMMENT (Nov. 17, 2022), https://www.gov.il/en/departments/news/most-news20221117; MIST, OFFICE OF LEGAL COUNSEL AND LEGISLATIVE AFFAIRS & MINISTRY OF JUSTICE, THE PRINCIPLES OF REGULATORY POLICY AND ETHICS IN THE FIELD OF ARTIFICIAL INTELLIGENCE (Oct. 30, 2022) (Isr.), https://www.gov.il/BlobFolder/rfp/061122/he/professional-letter.pdf.

In 2023, Israel issued "Israel's Policy on AI – Regulations and Ethics,"[80] which suggests (1) adopting sectoral regulation, (2) consistency with existing regulatory approach of leading countries and international organizations, (3) adopting a risk-based approach, (4) using soft regulatory tools intended to allow for an incremental development of the regulatory framework, and (5) fostering cooperation between the public and the private sectors.[81] It suggests creating an expert-based inter-agency body to guide sectoral regulators, enhance coordination, update governmental AI policies, advise AI regulation, and represent Israel globally.[82] Furthermore, it calls upon government bodies and regulators to identify and assess the deployment of AI systems and the associated challenges within their respective regulated sectors.[83]

### D. Switzerland: "Bottom-Up" and "Application-Focused" Approach

The Federal Council's Interdepartmental Working Group on AI, in its 2019 report titled "Challenges of Artificial Intelligence," pointed out that policy should ensure optimal, innovation-friendly frameworks that enable new technology development, leaving decisions regarding choosing specific technologies to individuals ("bottom-up" approach). [84] The report also noted that, because merely choosing AI among various technologies insufficiently justifies legal intervention, the impacts resulting from its application should be considered ("application-focused" approach). [85] The report further emphasized that the state can only intervene in the private domain where market failure occurs.[86] This has formed the basis of Switzerland's AI policy so far. After the E.U. AI Act was proposed, Swiss researchers at the University of Zurich proposed that Switzerland, through an adequately open legal framework,

---

[79] *Id.*
[80] MIST, ISRAEL'S POLICY ON ARTIFICIAL INTELLIGENCE – REGULATIONS AND ETHICS (DEC. 17, 2023), https://www.gov.il/en/pages/ai_2023.
[81] *Id.* at 3.
[82] *Id.* at 9.
[83] *Id.* at 9.
[84] Interdepartementale Arbeitsgruppe "Künstliche Intelligenz" an den Bundesrat, *Herausforderungen der künstlichen Intelligenz*, [Interdepartmental Working Group "Artificial Intelligence" to the Federal Council, *Challenges of Artificial Intelligence*] Dec. 13, 2019, at 34–37 (Switz.), https://www.sbfi.admin.ch/sbfi/de/home/bfi-politik/bfi-2021-2024/transversale-themen/digitalisierung-bfi/kuenstliche-intelligenz.html.
[85] *Id.*
[86] *Id.*

should provide the scope of activity (*Spielraum*) for Swiss companies not exporting to the E.U.[87]

However, because Switzerland is surrounded by the E.U., the country also seeks to mitigate the negative impacts of deviation from the E.U. framework. The Swiss Federal Department of Foreign Affairs (FDFA) pointed out that while Switzerland can decide on an independent legislative approach, major discrepancies from the E.U. framework may not align with Swiss supply chain interests.[88] The FDFA thus recommends: (1) establishing an advisory legal expert group; (2) coordinating Switzerland's international positions through the "Plateforme Tripartite" framework, a national information hub and multi-stakeholder platform for exchanging information on digital governance and AI; (3) strengthening cooperation with technical standards organizations; and (4) initiating negotiations with the E.U.[89] In November 2023, the Federal Council instructed the Federal Department of Environment, Transport, Energy, and Communications (DETEC) to prepare an overview of possible AI regulatory approaches, which is expected to be published by the end of 2024.[90]

### E. China: Sector-Specific Ordinances

China controls AI by implementing sector-specific ordinances based on existing laws, including the Cybersecurity Law, Data Security Law, Personal Information Protection Law, and Internet Information Services Management Measures. Notably, authorities like the Cyberspace Administration of China, the Ministry of Industry and Information Technology, the Ministry of Public Security, and the State Administration for Market Regulation have issued administrative regulations for different AI types. These include the Internet Information Service Algorithmic Recommendation Management Provisions,[91] the Internet Information Service Deep

---

[87] Digital Society Initiative, Universität Zürich [Zurich University], *Ein Rechtsrahmen für Künstliche Intelligenz* [A Legal Framework for Artificial Intelligence] (Nov. 2021) (Switz.), https://www.sachdokumentation.ch/bestand/ds/3728.

[88] SWISS FED. DEP'T. FOREIGN AFFS., ARTIFICIAL INTELLIGENCE AND INTERNATIONAL RULES – REPORT FOR THE FEDERAL COUNCIL 21 (Apr. 13, 2022) (Switz.), https://www.admin.ch/gov/en/start/documentation/media-releases.msg-id-88019.html.

[89] *Id.* at 21–23.

[90] SWISS FED. COUNCIL, FEDERAL COUNCIL EXAMINING REGULATORY APPROACHES TO AI (Nov. 22, 2023) (Switz.), https://www.admin.ch/gov/en/start/documentation/media-releases.msg-id-98791.html.

[91] Hulianwang Xinxi Fuwu Suanfa Tuijian Guanli Guiding [Internet Information Service Algorithmic Recommendation Management Provisions] (promulgated by the Cyberspace Admin. China et al., Dec. 31, 2021, effective Mar. 1, 2022) (China).

Synthesis Management Provisions,[92] and the Interim Management Measures for Generative AI Services.[93]

### III. STRIKING A BALANCE BETWEEN THE TWO APPROACHES

#### 1. Limitations of the Horizontal Approach

The E.U.'s horizontal approach reflects its commitment to safeguarding human rights, while avoiding regulatory vacuums. This approach is common in information technology law; digital transformation has driven industry convergence, exposed limitations and loopholes in traditional industry-specific regulations, and often required a horizontal approach for streamlining fragmented regulations.[94] However, AI-induced societal harms are not as uniform as the E.U. presumes; they are highly context-specific. Moreover, the degree of proportionality, granularity, and foreseeability embedded within the horizontal framework, particularly the E.U. AI Act, falls short of achieving global consensus.

#### A. Heterogeneity of AI-Induced Societal Challenges

Horizontal regulation schemes are prone to overregulation if they impose the same regulations on different subject matters. Crucially, AI-associated harms are less uniform than the E.U. assumes. The E.C.'s impact assessment for AI Regulation asserts that AI systems share five "problem drivers" and thus must be addressed by uniform regulations.[95] However, this argument is scientifically unsubstantiated.[96]

---

[92] Hulianwang Xinxi Fuwu Shendu Hecheng Guanli Guiding [Internet Information Service Deep Synthesis Management Provisions] (promulgated by the Cyberspace Admin. China et al., Nov. 25, 2022, effective Jan. 10, 2023) (China).
[93] Shengchengshi Rengongzhineng Fuwu Guanli Zanxingbanfa [Interim Measures for Management of Generative AI Service] (promulgated by the Cyberspace Admin. China et al., Jul. 10, 2023, effective Aug. 15, 2023) (China).
[94] For example, streamlining the uneven regulations of traditional broadcasting and over-the-top media into a single, horizontal regulatory regime has been regarded as the basis underpinning the reform of audiovisual media law.
[95] E.C., *supra* note 19, at 28–30.
[96] Opacity, complexity, autonomous behavior, continuous adaptation, and unpredictability are neither unique nor common to AI. Algorithms are not opaque, as discussed in Section IV. The opacity of models can constitute a concern for individuals classified or scored but not necessarily for those who interact with generative AI or autonomous AI. Complexity, the general nature of contemporary society, cannot justify creating unprecedented regulations. Autonomous behavior is a trait of autonomous AI justifying safety regulations but does not extend to other types of AI. Continuous adaptation that is vulnerable to data pollution attacks is a trait of continual learning, not AI in general. Human decision-making is often less predictable than algorithmic decision-making. Andrea Bonezzi et al., *The Human Black-Box: The Illusion of Understanding Human Better Than Algorithmic Decision-making*, 151(9) J. EXP. PSYCHOL. GEN. 2250 *passim* (2022). Data dependency has exceptional justification, as demonstrated by the failure of

In fact, traditional "AI ethics"—FAccT—are too contextual to generalize under a unitary regulatory framework. Allocative harms (disparate treatment or impact when allocating scarce resources, such as credit or employment opportunities) and representational harms (stereotyping, underrepresentation, or denigration harms) undermine fairness differently, requiring different solutions.[97] Infringement harms from copyright or privacy intrusions also require a distinct legal approach. Transparency and accountability are more elusive. The simple narrative that human autonomy should be defended against the threat of opaque "black box" AIs through mandated disclosure and audit inadequately explains what disclosure would need to be made, how it would be explained, or to what extent.[98] The genuine concerns and expectations of people interacting with AIs would be context-dependent. Those classified by hiring AIs or rated by credit-scoring AI would be curious about the rubrics. Individuals communicating with chatbots may want to know they are conversing with a non-human. Those closely interacting with autonomous AI, including passengers of driverless cars, might want test tracking, safety warning, and event data recording. One can liken the horizontal approach to chemotherapy, which is used to treat cancer *everywhere* in the body, and the context-specific approach to *targeted* radiation therapy.

If it is feasible to differentiate between harms arising from AI models and legislate them individually, pro-innovation jurisdictions would find it unsuitable to impose a one-size-fits-all bundle of regulations on any high-risk AI system. Section IV of this paper shows such categorization *is* feasible. Interestingly, with the rise of pretrained language models like ChatGPT, the E.U. has shifted from its horizontal framework and introduced vertical regulations specifically targeting the GPAI and generative AI.

### B. Insufficient Degree of Proportionality

A primary challenge from a bill attempting to apply uniform, comprehensive regulations to cover different subject matters is that some social benefits can be

---

autonomous AI or biases in hiring AI systems due to data incompleteness. However, the requirement that data be relevant, representative, error-free, and complete will impair data availability. Regulation is less likely to improve data quality than proactive policies that make more data available, such as deregulating the use of pseudonymized data, opening public databases, and ensuring the right to data portability.

[97] Kate Crawford, *The Trouble with Bias*, Neural Inf. Process. Syst. (NIPS) Keynote, YOUTUBE (2017), https://www.youtube.com/watch?v=fMym_BKWQzk.; Su Lin Blodgett et al., *Language (Technology) is Power: A Critical Survey of "Bias" in NLP*, ASSOC. COMPUT. LINGUISTICS (ACL), 5454–6 (2020).

[98] Mike Ananny & Kate Crawford, *Seeing Without Knowing: Limitations of the Transparency Ideal and its Application to Algorithmic Accountability*, 20 NEW MEDIA SOC. 1, 5–10 (2016).

attained through tailored, targeted regulations that carry smaller implementation costs. In other words, there is a potential for Pareto improvement: Costs can be reduced without sacrificing benefits by segregating regulations tailored for different AI.

Stringent provisions exacerbate the issue of insufficient proportionality, particularly for the E.U. and Brazil. Section II of this paper provided notable examples of stringent and burdensome provisions in the E.U. AI Act, such as technical documentation, the two-human monitoring requirement, and copyrighted training data disclosures. Nevertheless, the EC contends that the AI Act is "proportionate," calculating its total compliance cost at 30.5% of total high-risk AI investment cost[99] plus an upfront cost for a quality management system of up to €330,000 per firm.[100] This demonstrates several critical shortcomings. Firstly, it lacks justification for regulatory benefits exceeding 30.5% compliance costs. The E.C.'s impact assessment did not provide a concrete rationale for the 30.5% of high-risk AI investment allocated to regulatory costs, merely arguing the AI Act will address safety risks, fundamental rights risks, underenforcement, legal uncertainty, mistrust, and fragmentation,[101] and prevent the absence of regulation from impeding AI growth.[102] Secondly, the assessment disregards indirect costs, including the loss of investment and jobs. Contrary to the U.S.'s E.O. 12866, which includes "any adverse effects on the efficient functioning of the economy, private markets (including productivity, employment, and competitiveness), health, safety, and the natural environment"[103] in its cost considerations, the E.C.'s impact assessment ignored indirect costs. One estimation predicts the E.U. AI Act will reduce small and medium-sized enterprise profits by approximately 40%, impairing the E.U.'s growth potential and exacerbating market concentration.[104] Lastly, the E.C.'s impact assessment omits compliance costs arising from legal uncertainty due to overbroad and vague obligations and potential conflicts between different values. Accordingly, achieving consensus from jurisdictions that base their regulatory systems on rigorous cost-benefit analyses and proportionality may prove challenging.

---

[99] This includes 17% direct costs and 13.5% verification costs.
[100] ANDREA RENDA ET AL., STUDY TO SUPPORT AN IMPACT ASSESSMENT OF REGULATORY REQUIREMENTS FOR ARTIFICIAL INTELLIGENCE IN EUROPE, at 166 (2021), https://op.europa.eu/en/publication-detail/-/publication/55538b70-a638-11eb-9585-01aa75ed71a1.
[101] E.C., *supra* note 19, at 35–36.
[102] *Id.* at 38.
[103] Exec. Order No. 12866, Regulatory Planning and Review, 58 Fed. Reg. 51735 (1993), § 6(a)(3)(C)(ii).
[104] Benjamin Mueller, *How Much Will the Artificial Intelligence Act Cost Europe*?, CTR. FOR DATA INNOVATION 48 (July 26, 2021), https://datainnovation.org/2021/07/how-much-will-the-artificial-intelligence-act-cost-europe.

### C. Insufficient Degree of Granularity

Lack of proportionality is exacerbated by lack of granularity in classifying risks and defining the severity of legal obligations. The U.K.'s 2023 white paper highlights notable instances where low-risk AIs are classified as high risk under the E.U. AI Act: the "identification of superficial scratches on machinery" in critical infrastructure and a chatbot used by an online clothing retailer for customer service.[105] In these cases, the perceived harm associated with these AI applications can be significantly lower than the compliance burden imposed. The E.U. AI Act requires all high-risk AI systems to achieve an "appropriate level" of robustness throughout their lifecycle.[106] However, only some high-risk AI systems are deployed in unintended environments, and robustness may not be a serious and immediate legal concern unless related to safety or efficacy, as discussed in Section IV.

### D. Insufficient Degree of Foreseeability

Whether legal commands should be based on *ex ante* rules or *ex post* standards is contentiously debated in law and economics. Rules are costlier to create than standards but are more efficient if the law is frequently implemented. Whereas *ex post* standards require low upfront costs but high repeated implementation costs, *ex ante* rules require high one-time upfront costs but low implementation costs.[107] As AI evolves into a general-purpose technology that permeates ordinary life, a rule-based approach is sensible. Particularly when integrating trustworthy AI principles into the development process, converting these principles into metrics becomes crucial. Of course, the law must remain technology neutral, as mandating specific technologies or methodologies will lead to obsolescence. To that extent, adopting the principle-based approach by statute can be inevitable. Nevertheless, statutes should be supplemented by foreseeable metrics and standards outlined by "soft" laws.

---

[105] U.K. DSIT & OAI, *supra* note 15, at § 45.
[106] PE-CONS 24/24, art. 15(1).
[107] Louis Kaplow, *Rules versus Standards: An Economic Analysis*, 42 DUKE L. J. 557, 591–92, 621–22 (1992).

In fact, both the horizontal and context-specific approaches lean toward the principle-based approach. After releasing its first draft AI Act, the E.U. attempted to address criticism regarding the strictness and inflexibility of its rules by transitioning from a rule-based to a principle-based approach. However, this shift resulted in a tendency to overuse abstract adjectives and adverbs, [108] complicating legal definitions and triggering sanctions of up to 7% of worldwide revenue. To establish a coherent global norm, an initial upfront investment is required to clarify these terms before the world individually bears legal expenses and other costs for deciphering ambiguous terms.

### E. Potential Duplication and Conflict with Existing Regulations

AI systems categorized as "high risk" under an omnibus AI law are likely those employed in mission-critical domains already subject to stronger regulation, given their potential impact on the rights and interests of the public. In consequence, the horizontal framework can duplicate or conflict with existing sectoral regulations. For instance, a horizontal AI regulation ensuring fair access to loans for individuals may clash with traditional finance regulations aimed at balancing fair access with financial stability. Moreover, applying an omnibus AI law to medical AI could create conflict with existing medical device regulations such as the U.S. Food and Drug Administration's (FDA) oversight of AI/ML-based Software as Medical Devices (SaMDs), which establish distinct risk assessment and mitigation protocols.

### 2. Shaping the Context-Specific Approach

Given the horizontal approach's shortcomings, the context-specific approach holds greater promise. Nevertheless, further development regarding details, coherent regulatory objectives, and commensurable metrics are required for it to become an international norm.

Coherency will continue to be crucial in effectively coordinating fragmented regulations, both domestically and internationally. The U.K. government, while coining the term "coherent," emphasized formulating a "set of cross-sectoral principles" [109] and "central regulatory guidance" [110] to seamlessly coordinate domestic regulations. Inefficiency lies in imprudently selecting uniform, abstract principles. Thus, defining coherency should first begin by categorizing AIs based on distinct, concrete, and tangible societal harms.

---

[108] There are 284 uses of "relevant," 230 uses of "appropriate," 41 uses of "significant," 33 uses of "sufficient," 33 uses of "reasonable" or "reasonably" that exist in the May 14, 2024 text of the E.U. AI Act (PE-CONS 24/24).
[109] U.K. BEIS et al., *supra* note 20.
[110] U.K. DSIT & OAI, *supra* note 15, at § 73, Box 3.1.

To achieve a consistent and rational regulatory framework, debates regarding ethical "oughtness" must shift to a consideration of potential harms versus benefits. The regulatory regime should prioritize exploring commensurable metrics and applying them to measure and address each harm, instead of resorting to ill-defined principles. Trustworthy AI principles must be embedded into the system during development and converted into quantifiable metrics, as detailed in Section IV. The shortcomings of the E.C.'s initial proposal for the AI Act do not lie in its detailedness or concreteness, but in its overly stringent measures. Misunderstanding the problem has led both the E.U.'s amendment and the context-specific approach to derail into a principle-based approach that would impair foreseeability and commensurability without substantially improving proportionality. Such misunderstandings may result in additional compliance costs, including legal expenses to clarify the principles, and disperse the burden to a larger population that should have been subject to clearer rules.

The best way to enhance proportionality without compromising foreseeability is by complementing hard laws with soft laws, such as industry standards. While hard laws can establish regulatory objectives tailored to each AI category, soft laws provide metrics to formulate and achieve those objectives. In reaching this conclusion, this paper agrees with Timothy O'Reilly's view that AI fearmongering coupled with regulatory complexity can lead to analysis paralysis, and that developers should be encouraged to collaborate in establishing metrics and continuously updating them with emerging best practices.[111]

While a horizontal framework necessitates a centralized body to uniformly oversee AI systems, a contextual yet commensurable framework would be better suited for decentralized governance. In such a model, standards can evolve more dynamically and flexibly than rigid laws, occasionally competing with one another. Coherence can also play an important role in facilitating the interoperability and mutual recognition of these standards.

---

[111] Shiona McCallum & Chris Vallance, *ChatGPT-maker U-turns on Threat to Leave EU over AI law*, BBC (May 26, 2023), https://www.bbc.com/news/technology-65708114.

## IV. PROPOSING AN ALTERNATIVE 3C FRAMEWORK

### 1. Overview

This section proposes an alternative 3C framework addressing the limitations of the horizontal and context-specific approaches. To ensure *contextuality*, the 3C framework bifurcates the AI life cycle into two phases: (1) learning; and (2) deployment for specific tasks, which are categorized into autonomous, allocative, punitive, cognitive, and generative AI. To ensure *coherency* within a category, each category of "tasks" is assigned specific regulatory objectives: safety for autonomous AI; fairness and explainability for allocative AI; accuracy and explainability for punitive AI; accuracy, robustness, and privacy for cognitive AI; and the mitigation of infringement and misuse for generative AI. The framework further addresses the learning process with a focus on optimizing liability related to data mining, copyright, and personal data. To ensure *commensurability* regarding each regulatory objective, the framework promotes adopting international industry standards that convert principles into quantifiable metrics.

### 2. *Contextuality*: Categorizing AI

Rather than define and separately govern foundation models or GPAI, the 3C framework focuses on the learning phase and derives social problems arising from it before categorizing tasks. The societal harms vary sufficiently among autonomous, discriminative, and generative tasks to warrant distinct legal treatments. Notably, there is overlap among the AI types. Autonomous AI sensors fall under cognitive AI. Transformer-based models, which perform generative tasks by themselves, can be fine-tuned to implement downstream tasks falling under allocative, punitive, or cognitive AI. Humanoids can classify people, speak, or paint. In such cases, regulations should be applied based on the specific tasks performed by the AI system.

#### A. Learning Phase

##### (1) Training a Limited-Purpose Model

A limited-purpose model encompasses traditional supervised, unsupervised, or reinforcement learning models that are typically trained from the ground-up for specific tasks.

##### (2) Training and Adapting a General-Purpose Model

General-purpose models are "trained on broad data (generally using self-supervision at scale) that can be adapted (e.g., fine-tuned) to a wide range of downstream tasks."[112] The U.S., the E.U., and the U.K. refer to a general-purpose model as a foundation model,[113] GPAI model,[114] and (if highly capable) frontier AI,[115] respectively.

The Transformer architecture enabled the training of general-purpose models. Transformer is a neural network designed for processing sequential data like text, image, and voice.[116] Its key innovation is the multi-head self-attention mechanism, which allows the model to focus on different parts of the input sequence in parallel.[117] This mechanism enables the model to capture contextual relationships between tokens (such as words, word morphemes, and image particles) in the sequence,[118] making it effective for tasks like dialogue, translation, and summarization. Google's Bidirectional Encoder Representations from Transformers (BERT) and OpenAI's Generative Pre-Trained Transformer (GPT) were developed from encoder and decoder, respectively, in the Transformer structure.[119]

Transformer-based pre-trained models, including ChatGPT, can be utilized for generative tasks without additional adaptations. However, what characterizes

---

[112] Rishi Bommasani et al., *On the Opportunities and Risks of Foundation Models*, CTR. FOR RSCH. ON FOUND. MODELS (CFRM), July 12, 2022, at 3, https://doi.org/10.48550/arXiv.2108.07258.

[113] E.O. 14110 § 3(k) defines a "dual-use foundation model" as an "AI model that is trained on broad data; generally uses self-supervision; contains at least tens of billions of parameters; is applicable across a wide range of contexts; and that exhibits, or could be easily modified to exhibit, high levels of performance at tasks that pose a serious risk to security, national economic security, national public health or safety, or any combination of those matters, such as by: (1) substantially lowering the barrier of entry for non-experts to design, synthesize, acquire, or use CBRN weapons; (2) enabling powerful offensive cyber operations through automated vulnerability discovery and exploitation against a wide range of potential targets of cyber attacks; or (3) permitting the evasion of human control or oversight through means of deception or obfuscation."

[114] The EU AI Act defines a GPAI model as an "AI model, including where such an AI model is trained with a large amount of data using self-supervision at scale, that displays significant generality and is capable of competently performing a wide range of distinct tasks regardless of the way the model is placed on the market and that can be integrated into a variety of downstream systems or applications," and excludes "AI models that are used for research, development or prototyping activities before they are placed on the market." Art. 3(63).

[115] The U.K. defines "frontier AI" as "highly capable general-purpose AI models that can perform a wide variety of tasks and match or exceed the capabilities present in today's most advanced models." U.K. DSIT, *supra* note 13, at 4.

[116] Ashish Vaswani et al., *Attention Is All You Need*, 30 ADVANCES NEURAL INF. PROCESS. SYST. 3023 *passim* (2017).

[117] *Id.*

[118] *Id.*

[119] Jacob Devlin et al., *BERT: Pre-training of Deep Bidirectional Transformers for Language Understanding*, ASSOC. COMPUT. LINGUISTICS. (ACL) 4171, 4173–74, n. 4 (2019).

Transformer-based models as general-purpose models is their ability to serve diverse downstream tasks through the adaptations of pre-trained models, such as fine-tuning and in-context learning. Through "fine-tuning," a pre-trained model is retrained on a specific downstream task by adjusting its parameters with additional data tailored for that task.[120] This process allows the model to adapt its learned representations to better suit the target task. In addition, GPT-3 and later AI models adapt to specific downstream tasks by temporarily learning from prompts—"in-context learning" or "prompt engineering"—without modifying parameters through fine-tuning.[121] The models implement "few-shot learning," meaning they can be prompted to generalize and generate appropriate responses for similar tasks.[122] For instance, a ChatGPT user inserting prompts such as "U.S. => D.C., India => New Delhi, South Korea =>" to obtain an answer "Seoul."

In addition to Transformer-based decoders (such as GPT, Llama, and Gemini) or encoders (such as BERT), the Contrastive Language-Image Pre-training (CLIP) is arguably a general-purpose model.[123] CLIP models, commonly used for text-to-image (T2I) or text-to-video (T2V) models, are pre-trained using large datasets containing images and corresponding textual descriptions to align the visual and textual representations in a shared vector space.[124]

### B. Task Phase

#### (1) Autonomous AI

Robots and other autonomous AIs perceive environment, reason, and control actuators. They include self-driving vehicles (e.g. autos, aircraft, seacraft, trains, cableways, amusement park rides, and elevators); self-operating facilities (e.g. nuclear plants, industrial robots, electricity facilities, gas stores, and dams); surgical robots; and pricing agents (e.g. algorithmic trading agents). Autonomous AI uses cognitive AI for perceiving environments and uses discriminative or generative AI for reasoning. However, they should be situated differently in the legal system given the safety risks, particularly when self-operating in a public space.

---

[120] *Id.* at 4172–73, 4175. Fine-tuning should be distinguished from domain adaptive learning, which feeds additional data in the target domain to a model that has been pretrained in the source domain. Shai Ben-David et al., *A Theory of Learning from Different Domains*, 79 MACH. LEARN. 151, 154 (2010).

[121] Tom B. Brown et al., *Language Models are Few-Shot Learners*, 33 ADVANCES IN NEURAL INF. PROCESS. SYST. 1877, 1877–1901 (2020).

[122] *Id.*

[123] Bommasani et al., *supra* note 112, at 3.

[124] Alec Radford et al., *Learning Transferable Visual Models from Natural Language Supervision*, INT. CONF. MAC. LEARN. (ICML) 139 *passim* (2021).

### (2) Discriminative AI

Discriminative AI scores or classifies people to assign benefits or sanctions or identifies persons, their states, or objects from the dataset. Models are often built from labeled data through supervised learning. However, pre-trained transformer models can also be fine-tuned to perform the discriminative task. Discriminative AI can be classified into the following subcategories according to how they interact with and affect people:

- *Allocative AI*: "Allocative AI" is used to allocate a limited amount of goods, including resources or opportunities, to individuals. It encompasses various applications, including AI-powered hiring, admission processes, credit scoring, insurance underwriting, and ranking systems. Though AI can theoretically be applied to allocate unlimited goods, including unlimited virtual assets, such occurrences are highly infrequent. Thus, this paper focuses on the allocation of scarce goods. In such case, allocative AI replaces or complements *human relative evaluation*, including grading on a curve. As such, its discrimination and inexplicability must be scrutinized.

- *Punitive AI*: "Punitive AI" imposes negative consequences, particularly sanctions, on individuals. This category includes AI-powered applications in criminal sentencing, fraud detection, claim adjusting, and immigration control. While AI could theoretically assign a limited amount of sanction, with jail capacity overflowing or akin to the historical practice of decimation in ancient Rome, such instances remain infrequent. Thus, limited sanctions are outside the scope of this paper. Punitive AI thus replaces or complements *human absolute evaluation*; inaccuracy and inexplicability must be scrutinized.

- *Cognitive AI*: "Cognitive AI" is used to perceive or recognize fact. It includes computer vision, AI imaging and diagnostics, biometric identification, and recommendation. Cognitive AI can replace or complement *human cognition*; thus, QoS problems and disparity must be scrutinized, especially when related to safety or efficacy, such as for autonomous AI sensors and medical AI sensors. Additionally, privacy invasion, including mass surveillance risks, are an important cognitive AI concern.

### (3) Generative AI

E.O. 14110 defines generative AI as "the class of AI models that emulate the structure and characteristics of input data in order to generate derived synthetic content."[125] Generative AI fuels AI-powered writing, compositions, painting, machine translations, image captioning, and other tasks. As noted, Transformer-based decoders, including GPT, Llama, and Gemini, can be used for generative tasks without adaptations. Variational auto-encoder (VAE),[126] GAN,[127] and diffusion models[128] provide more examples of generative AI with powerful decoding capabilities. Currently, most T2I models, including DALL·E 2 or 3, or Stable Diffusion, retrieve visual representations semantically related to text prompts through CLIP and generate new images from the representations using diffusion.[129]

### 3. *Coherency*: Assigning Key Regulatory Objectives to Each Phase/Category

#### A. Learning Phase

##### (1) Special Regulations over High-Performance General-Purpose Models

Stanford Human-Centered AI researchers issued a report warning that the "foundation model" will attain an unprecedented level of "homogenization" or "algorithmic monoculture" across various downstream tasks, creating "single points of failure" and amplifying biases.[130] This conjecture sparked debates regarding implementing special regulations over general-purpose models. Monocultures, widespread adoption of a single model or technology, render social systems vulnerable to attacks that exploit common patterns.[131] However, further evidence is needed to broaden this argument to encompass other domains such as bias, safety, and infringement issues.

---

[125] 88 Fed. Reg. 75191, § 3(p).
[126] Diederik P. Kingma & Max Welling, *Auto-Encoding Variational Bayes*, INT'L CONF. LEARN. REPRESENTATIONS (ICLR) *passim* (2014).
[127] Ian J. Goodfellow et al., *Generative Adversarial Networks*, 63 COMM. ACM 139 *passim* (2020).
[128] Jonathan Ho et al., *Denoising Diffusion Probabilistic Models*, 33 ADVANCES NEURAL INF. PROCESS. SYST. 6840, 6840–51 (2020).
[129] Aditya Ramesh et al., *Hierarchical Text-Conditional Image Generation with CLIP Latents*, ARXIV:2204.06125 *passim* (Apr. 13, 2022), https://arxiv.org/abs/2204.06125; Robin Rombach et al., *High-Resolution Image Synthesis with Latent Diffusion Models*, COMPUT. VIS. PATTERN RECOGNIT. (CVPR) *passim* (2022).
[130] Bommasani et al., *supra* note 112 *passim*.
[131] Randal C. Picker, *Cybersecurity: Of Heterogeneity and Autarky* (The Law and Economics of Cybersecurity No. 223, 2004), at 115, 119–30 (2005).

That said, special regulations *have* been devised to control high-performance general-purpose models. As noted in Section II, in the U.S., E.O. 14110 delineates a "dual-use foundation model," defined based on a threshold of 10 billion parameters, and aims to address security, national security, and public health or safety, to implement security control over it.[132] The E.U. AI Act requires providers of GPAI models with "systemic risks," which is presumed when the cumulative amount of computation used for training measured in floating point operations (FLOPs) exceeds $10^{25}$,[133] to perform model evaluations, assess and mitigate systemic risks, report serious incidents, and ensure adequate cybersecurity protections.[134]

From a national security perspective, it would be inevitable to define a general-purpose model, as exemplified in E.O. 14110. Without a clear definition, it will not be possible to specify the target of export control or other measures concerning the global value chain. For other regulatory objectives, however, it would be more prudent to focus on specific tasks, as harm often emerges at the intersection of people and AI rather than during the learning process, until further evidence is gathered. In decentralized governance, AISIs are expected to play a pivotal role in collecting evidence related to unforeseen risks linked with general-purpose models and refining tools for measuring and mitigating these risks.

### (2) Transparency Requirement

The E.U. AI Act requires GPAI model providers to (1) keep technical documentation, (2) provide it to downstream providers, (3) implement a copyright policy, and (4) disclose a summary of the training content used.[135] The mandatory disclosure requirement causes significant implementation costs but is unlikely to effectively mitigate safety risks.[136] In fact, mandatory disclosure may cause unintended harms, including crowding out important information and undermining consumer protections.[137] Mandatory disclosure might become a mere façade for developers who do not really invest in mitigating technologies. Furthermore, an overly detailed description of models in technical documentation could facilitate adversarial attacks. As noted, training data disclosure is impractical due to the enormous volume of training datasets involved.

---

[132] 88 Fed. Reg. 75191, §§ 3(k), 4.2(a).
[133] PE-CONS 24/24, art. 51(2). FLOPs mean any "mathematical operation or assignment involving floating-point numbers, which are a subset of the real numbers typically represented on computers by an integer of fixed precision scaled by an integer exponent of a fixed base." Art. 3(67).
[134] Art. 55(1).
[135] Art. 53(1).

In contrast, E.O. 14110 scrutinizes the widely available model weights (or parameters) of dual-use foundation models, which are often accessible through open-source licensing. This stance might stem from concerns that such practices could hinder the efficacy of export control measures over the models. The DOC is tasked with reporting to the President regarding the implications of such dual-use foundation models with widely available model weights after soliciting input from stakeholders through public consultation.[138] This has the potential for impacting the future trajectory of AI, turning open-source models such as Llama into closed systems.

### (3) Optimizing Liability for Potential Infringement During Training

Many jurisdictions have debated whether data mining—including web scraping or crawling—and model training breach dataset rights, copyright, or personal data regulations. This paper argues that to prevent existing regulations from disproportionately impeding the "harvest of knowledge,"[139] in which social benefits surpass costs, the legal system should actively embrace existing safe harbors, including the fair use doctrine; text and data mining (TDM) exceptions; and the compatible use of personal data. The issue of potential infringement during data mining and model training is distinct from the issue of infringement during generation, which will be separately addressed when discussing generative AI.

#### *a. Dataset Rights*

---

[136] Omri Ben-Shahar & Carl E. Schneider, *The Failure of Mandated Disclosure*, 159 U. PA. L. REV. 647, 735–37 (2011).
[137] *Id.* at 737–42.
[138] 88 Fed. Reg. 75191, § 4.6.
[139] Harper & Row, Publishers, Inc. v. Nation Enterprises, 471 U.S. 539, 545 (1985).

Web scraping may infringe database rights held by website administrators in jurisdictions where *sui generis* database rights are recognized, including the E.U., Mexico, and South Korea.[140] In the U.S., legal risks involving web scraping have decreased since the Ninth Circuit's 2019 decision in *hiQ Labs v. LinkedIn*[141]. There, the court affirmed the district court's award of a preliminary injection in hiQ's favor, holding that LinkedIn's selective ban of hiQ's access to member profiles (1) did not constitute unauthorized access to LinkedIn's server computers under the Computer Fraud and Abuse Act[142] and (2) raised serious questions on the merits of claims for unfair competition under California's Unfair Competition Law[143].[144] Despite this ruling, applying trespassing doctrine to web scraping remains contentious in the U.S.[145] Dataset rights outside of the U.S. are grounded in the "sweat of the brow" doctrine, which the U.S. Supreme Court rejected in its 1991 decision, *Feist v. Rural*.[146] Dataset rights must be reassessed given the ongoing reduction in data collection costs and the shift in focus from labeled data to unstructured corpora.

### b. Copyright

Determining "substantial transformative use" within the fair use doctrine is crucial to U.S. court precedent.[147] Though data mining can entail duplicating or transmitting copyrighted materials, some could recognize it as "transformative use." For example, the learning phase fits parameters that do not directly contain express components unlike generative AI, where decoding often necessitates the duplication of tokens which may encompass expressive elements. Other jurisdictions have attempted to mitigate developers' liability for copyright infringement during TDM.[148] However, the scope of the TDM exception is limited in many jurisdictions, as it often requires non-commercial use.

---

[140] Commission and Parliament Directive 1996/9/EC, On the Legal Protection of Databases, 1996 O.J. (L 77) 20; Ley Federal del Derecho de Autor 24-12-1996 [Federal Copyright Law], arts. 104, 107–10, Diario Oficial de la Federación [DOF], últimas reformas DOF 1-7-2020 [Official Gazette of the Federation [DOF], latest reforms DOF 1-7-2020] (Mex.); Jeojakgweonbeob [Copyright Act], Act No. 432, Jan. 28, 1957, *amended* by Act No. 20358, Feb. 27, 2024, art. 93(1) (S. Kor.).
[141] HiQ Labs, Inc. v. LinkedIn Corp., 938 F.3d 985 (9th Cir. 2019), 141 S. Ct. 2752, 210 L. Ed. 2d 902 (2021), 31 F.4th 1180 (9th Cir. 2022).
[142] 18 U.S.C. § 1030.
[143] CAL. BUS. & PROF. CODE §§ 17200.
[144] HiQ Labs, 938 F.3d, at 995–1004.
[145] *See, e.g.*, eBay v. Bidder's Edge, 100 F. Supp. 2d 1058 (N.D. Cal. 2000) (granting eBay a preliminary injunction against an auction data aggregator's crawling based the trespass to chattels doctrine).
[146] Feist Publications, Inc., v. Rural Telephone Service Co., 499 U.S. 340 (1991).
[147] *See, e.g.,* Authors Guild v. HathiTrust, 755 F.3d 87 (2d Cir. 2014); Google LLC v. Oracle Am., Inc., 593 U.S. __ (2021).

### c. Personal Data

Issues vary according to the sources of data, including (1) publicly available sources; (2) users, such as user prompts; and (3) third parties. Regarding publicly available data, certain jurisdictions exempt such data from consent requirements under data protection laws.[149] The CPRA recently amended the California Consumer Privacy Act of 2018 (CCPA) so as to widen the scope of publicly available data (excluding biometric data) which is exempted from protected personal data.[150] Virginia[151] and Utah[152] enacted similar statutes, while Colorado[153] and Connecticut[154] have enacted less lenient statutes.

In other cases, jurisdictions, particularly those who modeled their data protection laws after the GDPR, take issue with whether the processing meets data protection law requirements, including consent, legitimate interest, compatible use, or pseudonymization. In 2021, South Korea's Personal Information Protection Commission (PIPC) issued a decision requiring pseudonymization for training a chatbot service.[155] In 2023, Italy's data protection agency, *Garante*, issued a cease-processing order against ChatGPT due to its potential violation of GDPR.[156]

---

[148] Chosakukenhō [Copyright Act], Law No. 30 of 2018 (Japan), art. 30-4 and 47-5(1)(ii); Directive 2019/790 of the European Parliament and of the Council of 17 April 2019 on Copyright and Related Rights in the Digital Single Market and Amending Directives 96/9/EC and 2001/29/EC, PE/51/2019/REV/1, 2019 O.J. (L 130) 92–125 (EU); Copyright, Designs and Patents Act 1988, 1988 c. 48, § 29A, as amended by the Copyright and Rights in Performances (Research, Education, Libraries and Archives) Regulations 2014, SI 2014/1372, art. 2 (U.K.); Code de la propriété intellectuelle [INTELLECTUAL PROPERTY CODE], art. L122-5, -10,L-342-3, -5 (2016), as amended by Loi 2016-1321 du 7 octobre 2016 pour une République numérique [Law 2016-1321 of Oct. 7, 2016 for the Digital Republic] (Fr.); Gesetz über Urheberrecht und verwandte Schutzrechte [URHG] [LAW ON COPYRIGHT AND RELATED RIGHTS], § 60d, as amended by Gesetz zur Angleichung des Urheberrechts an die aktuellen Erfordernisse der Wissensgesellschaft [URHWISSG][Law Adjusting Copyright to the Current Needs of the Knowledge Society], Sept. 1, 2017 (Ger.).

[149] *See, e.g.*, Personal Information Protection and Electronic Documents Act, S.C. 2000, c 5, art. 7(1)(d), (2)(c.1), (3)(h.1) (Can.); Personal Data Protection Act 2012, sec. 17(1), First Schedule, Part 2(1) (Singapore).

[150] CAL. CIV. CODE § 1798.140(v)(2) (2022).

[151] VA. CODE ANN. § 59.1-575 (2021).

[152] UTAH CODE ANN. § 13-61-101(29) (2022).

[153] COLO. REV. STAT. § 6-1-1303(17)(b) (2021).

[154] CONN. SUB. BILL No. 6, § 1(25) (2022).

[155] Personal Information Protection Commission, Decision No. 2021-007-072, Apr. 28, 2021.

[156] In March 2023, Italy's data protection agency, Garante, issued a cease-processing order against ChatGPT due to its potential violation of GDPR. Garante Per La Protezione Dei Dati Personali [GPDP][Italian Data Protection Authority], *Artificial Intelligence: Stop to ChatGPT by the Italian SA – Personal Data is Collected Unlawfully, No*

Notably, the E.U. AI Act prohibits building facial recognition databases through untargeted scraping from the Internet or closed-circuit television.[157]

## B. Autonomous AI

### (1) Adapting Safety Regulations

Autonomous AI has led to accidents, including self-driving car and drone accidents. Thus, the legal system must address this safety deficiency. While AI ethics have emphasized transparency and fairness, safety has not attracted the attention it deserves. Ironically, this lack of emphasis on safety has steered clear of unnecessary attempts at horizontal regulation and has facilitated more sound sectoral approaches. An example of this is the swift adaptation of vehicle safety regulations to self-driving cars.

Existing safety regulations, including those for vehicles (e.g. autos, aircraft, elevators) and facilities (e.g., nuclear plants, industrial facilities, electricity facilities) must be adapted to address incremental safety risks arising from embedded autonomous AI. One adaptation is mandating a kill switch to ensure human control. Self-driving car regulations, which have already been prepared and implemented worldwide,[158] can provide a model for this pan-industry adjustment.

Security issues are closely related to the safety of autonomous AI systems. For instance, if a self-driving car is hacked, it can be exploited for criminal activities. Therefore, addressing these security concerns is crucial for safety regulations. In 2020, the United Nations introduced UN Regulation No. 155, which governs the implementation of a cybersecurity management system applicable to wheeled

---

*Age Verification System is in Place for Children* (Mar. 31, 2023), https://www.garanteprivacy.it/home/docweb/-/docweb-display/docweb/9870847. OpenAI promptly restricted access to ChatGPT from within Italy thereafter. In April 2023, Garante outlined several measures that OpenAI must undertake before the order against ChatGPT is rescinded: (1) publish a notice elucidating the processes and rationale behind ChatGPT's data handling procedures and the rights granted to individuals, (2) eliminate references to contractual obligations and instead establish either consent or legitimate interest as the legal basis for data processing, (3) ensure that individuals possess the means to exercise their rights to correct or erase their personal data, and (4) implement an age verification system to safeguard minors' data and privacy. Garante Per La Protezione Dei Dati Personali, *ChatGPT: Italian SA to lift temporary limitation if OpenAI implements measures – 30 April set as deadline for compliance* (Apr. 21, 2023), https://www.garanteprivacy.it/home/docweb/-/docweb-display/docweb/9874751. In response to these requirements, OpenAI augmented privacy disclosures and controls, lifting geo-blocking in Italy on April 29, 2023.

[157] PE-CONS 24/24, art. 5(1)(e).
[158] See, for example, amendments to the 1968 Vienna Convention of Road Traffic, Nov. 8, 1968, 1912 U.N.T.S. 295, the United Nations Regulations No. 155 to 157, and the Federal Motor Vehicle Safety Standards, 49 C.F.R. § 571 (2024).

vehicles, demonstrating the increasing importance of cybersecurity measures in autonomous technologies.[159]

### (2) Transparency from the Perspective of Autonomous AI: Test Tracking, Safety Warning, and Event Data Recording

To address incremental safety risks arising from autonomous robots, including self-driving vehicles, a different notion of transparency must be ensured at different stages: (1) before accidents, test tracking; (2) for impending accidents, safety warning; and (3) after accidents, event data recording. Under the U.S. National Highway Traffic Safety Administration (NHTSA)'s Automated Vehicle Transparency and Engagement for Safe Testing (AV TEST) Initiative, states and companies have voluntarily disclosed testing information to the NHTSA.[160] For Level 3 self-driving cars (those with conditional automation), a pending issue regarding safety is warning drivers so they can take timely control of wheels and brakes upon emergency.[161] Another issue is whether and how to provide safety information to other road users.[162] Though South Korea currently requires Level 3 vehicles to carry signage announcing they are self-driving,[163] its justification will wane as driverless cars become prevalent. The mandatory installation of event data recorders (EDRs), informally dubbed "black boxes," can play a crucial role in ensuring traceability and evidence collection and deepen our understanding of autonomous AI safety problems. Starting from July 6, 2022, the E.U. required all automated vehicles to be equipped with EDRs, defined as systems "with the only purpose of recording and storing critical crash-related parameters and information shortly before, during and immediately after a collision."[164]

---

[159] UN Regulation No. 15–5—Cyber security and cyber security management system, Apr. 3, 2021, E/ECE/TRANS/505/Rev.3/Add.154.

[160] Press Release, NHTSA, U.S. Transportation Secretary Elaine L. Chao Announces First Participants in New Automated Vehicle Initiative Web Pilot to Improve Safety, Testing, Public Engagement (Jun. 15, 2020), https://www.nhtsa.gov/press-releases/us-transportation-secretary-elaine-l-chao-announces-first-participants-new-automated.

[161] Taxonomy and Definitions for Terms Related to Driving Automation Systems for On-Road Motor Vehicles No. J3016_202104 (SAE INT'L 2021), https://www.sae.org/standards/content/j3016_202104.

[162] Regulation (EU) 2019/2144 of the European Parliament and of the Council of 27 November 2019 on Type-approval Requirements for Motor Vehicles and Their Trailers, and Systems, Components and Separate Technical Units Intended for Such Vehicles, as Regards Their General Safety and the Protection of Vehicle Occupants and Vulnerable Road Users. Commission Regulation 2019/2144, art. 11(1)(f), 2019 O.J. (L 325) 14, 1–40.

[163] Jadongcha Gwanri-beop Shihaeng-gyuchik [Enforcement Regulation for the Motor Vehicle Management Act], MOLIT Regulation No. 861, Aug. 1, 1987, *amended* by Regulation No. 1155, Oct. 26, 2022, art. 26-2(1)(v) (S. Kor.).

[164] Regulation (EU) 2019/2144, arts. 3(13), 11(1)(d).

### (3) Pricing Agents

Potential economic harm from automated pricing agents acting in the market requires sector-specific review. What causes algorithmic trading agents to amplify systemic risks and volatility in stock markets, as observed in the 2010 Flash Crash,[165] and whether algorithmic tacit-collusion conjecture[166] has empirical support, remain disputed.

### (4) Fairness and Other Principles

Proponents of the context-specific approach believe that while safe self-driving elevators or cars are required, it crosses the line to force "transparent" self-driving elevators or "fair" self-driving cars merely because AI is used. Horizontal approach proponents argue that non-discrimination regulations should still apply to all self-driving cars. For example, self-driving cars are biased toward identifying pedestrians with light skin tones more accurately than those with darker skin tones.[167] This paper takes such problems seriously. However, it classifies computer vision in self-driving cars as a component of cognitive AI, rather than autonomous AI itself, and proposes solutions tailored to the QoS disparity, as discussed below.

## C. Allocative AI

### (1) Addressing Discrimination

It is well-documented that allocative AI's biased scoring or classification can unfairly allocate resources or opportunities.[168] As noted, fairness issues include (1) allocative harms, wherein a model allocates resources, and (2) representational harms, including stereotyping, underrepresentation, and denigration.[169] While allocative harms are "immediate, easily quantifiable, and discrete," representational harms are "long-term, difficult to formalize, and diffuse."[170]

---

[165] *See generally,* Andrei Kirilenko et al., *The Flash Crash: High-Frequency Trading in an Electronic Market*, 72 J. FIN. 967 (2017).
[166] *See generally,* ARIEL EZRACHI & MAURICE E. STUCKE, VIRTUAL COMPETITION: THE PROMISE AND PERILS OF THE ALGORITHM-DRIVEN ECONOMY (2016).
[167] *See generally,* Benjamin Wilson, Judy Hoffman & Jamie Morgenstern, *Predictive Inequality in Object Detection*, ARXIV:1902.11097 (Feb. 21, 2019), https://arxiv.org/pdf/1902.11097.
[168] *See generally,* Mikella Hurley & Julius Adebayo, *Credit Scoring in the Era of Big Data*, 18 YALE J. L. TECH. 156 (2016) (discussing potential discrimination in big-data credit scoring tools); Talia B. Gillis & Jann L. Spiess, *Big Data and Discrimination*, 86 U. CHI. L. REV. 459 (2019) (discussing potential discrimination in big-data mortgage lending tools).
[169] Crawford, *supra* note 97.

One critical allocative AI issue is addressing allocative harms. Discriminative AI—especially cognitive AI—can also cause representational harms, including denigration. Such was the case when Google Photo misidentified a black couple as primates in 2015.[171] Underrepresentation poses another allocative harm. For example, gender disparities can lead to inaccurate diagnoses. However, the impact of such representational harms should be examined from a long-term viewpoint while acknowledging that allocative harms might call for immediate diagnosis and correction.

There are several questions for an international norm to answer to systemize allocative fairness inquiries. First, how should a system measure disparity? There is a consensus regarding equality of opportunity, though equality of outcome can be appropriately considered where serious inequalities exist. Equality of opportunity is mathematically interpreted as a parity of accuracy (or error) rates. Individuals with the same qualifications are expected to have an equal chance of passing or failing, regardless of sensitive attributes like gender and race. Paragraph 4 of this section provides further discussion.

Second, how would a system define sensitive attributes or protected groups?[172] Defining sensitive attributes is a normative question. The U.S. Supreme Court currently categorizes classifications into (1) "suspect" classifications requiring strict scrutiny, such as race[173]; (2) classifications requiring relatively reduced scrutiny, such as gender,[174] illegitimacy,[175] and alienage[176]; and (3) other classifications subject to minimum rationality standards, such as age[177] and wealth.[178] Additionally, non-discrimination statutes define different sensitive attributes. For example, Title VII of the Civil Rights Act of 1964 bans discrimination of employees on account of their race, color, region, sex, or national origin.[179]

---

[170] *Id.*
[171] Conor Dougherty, *Google Photo Mistakenly Labels Black People 'Gorillas'*, N.Y. TIMES, Jul. 7, 2015, https://archive.nytimes.com/bits.blogs.nytimes.com/2015/07/01/google-photos-mistakenly-labels-black-people-gorillas.
[172] Non-discrimination differs from non-differentiation (such as a ban on differential pricing under the Robinson-Patman Act), although the literature tends to conflate these. The former prohibits disparate treatment of or disparate impact over a specific protected group based on sensitive attributes such as race and gender. The latter pursues uniform outcomes (such as the Law of One Price) regardless of attributes. AI should be allowed to classify individuals unless the AI unfairly classifies protected groups because of sensitive attributes.
[173] *See* Korematsu v. United States, 323 U.S. 214, 216 (1944).

Third, would a system adopt an input- or output-centered approach? The input-centric approach removes a feature representing sensitive attributes from a feature vector, while the output-centric approach further examines disparity in predictions between the protected group and others.[180] The input-centric approach can help forestall disparate treatment, while the output-centric approach typically aims at avoiding disparate impact, although these computer science concepts (input-centric v. output-centric) and legal concepts (disparate treatment v. disparate impact) do not exactly match.[181] The disparate impact theory was recognized by the U.S. Supreme Court's 1971 decision in *Griggs v. Duke Power Co.*, as applied to Title VII of the Civil Rights Act of 1964,[182] and was enshrined in the disparate-impact provisions of Title VII,[183] the Age Discrimination in Employment Act of 1967,[184] and Title VIII of the Civil Rights Act of 1968 (the Fair Housing Act of 1968).[185] In these areas requiring disparate impact, the output-centric approach can be more reasonable. However, a disparate-impact claim is not cognizable in other areas, including the

---

[174] *See* Craig v. Boren, 429 U.S. 190, 197 (1976); Califano v. Goldfarb, 430 U.S. 199, 210–11 (1977); Califano v. Webster, 430 U.S. 313, 316–317 (1977); Orr v. Orr, 440 U.S. 268, 279 (1979); Caban v. Mohammed, 441 U.S. 380, 388 (1979); Personnel Admin. Mass. v. Feeney, 442 U.S. 256, 273 (1979); Califano v. Westcott, 443 U.S. 76, 85 (1979); Wengler v. Druggists Mut. Ins. Co., 446 U.S. 142, 150 (1980); Kirchberg v. Feenstra, 450 U.S. 455, 461 (1981); Miss. Univ. for Women v. Hogan, 458 U.S. 718, 723–24 (1982).

[175] Mills v. Habluetzel, 456 U.S. 91, 99 (1982); Parham v. Hughes, 441 U.S. 347 (1979); Lalli v. Lalli, 439 U.S. 259 (1978); Trimble v. Gordon, 430 U.S. 762 (1977).

[176] Plyler v. Doe, 457 U.S. 202 (1982).

[177] Mass. Bd. Retirement v. Murgia, 427 U.S. 307 (1976); Vance v. Bradley, 440 U.S. 93 (1979); Gregory v. Ashcroft, 501 U.S. 452, 470 (1991).

[178] San Antonio School Dist. v. Rodriguez, 411 U.S. 1, 2 (1973). United States v. Kras, 409 U.S. 434, 446 (1973); Maher v. Roe, 432 U.S. 464, 470 (1977); Harris v. McRae, 448 U.S. 297, 324 (1980).

[179] 42 U.S.C. § 2000e–2.

[180] Gillis & Spiess, *supra* note 1688, at 467–73, 479–87.

[181] *But see id.* at 460–62 (exploring input-centric and output-centric approaches from a distinct perspective compared to the analysis of disparate treatment versus disparate impact).

[182] Griggs v. Duke Power Co., 401 U.S. 424 (1971).

[183] 78 Stat. 255.

[184] 29 U.S.C. §§ 621–34.

[185] 7 CFR § 1901.203. *See* Tex. Dept. Hous. & Cmty. Affs. v. Inclusive Cmtys. Project, Inc., 576 U.S. 519 (2015).

equal protection clauses of the Fifth and Fourteenth Amendments of the U.S. Constitution.[186] Some argue the output-centric approach must be prioritized in machine learning (ML) settings because the input-centric approach is prone to a proxy problem[187]: Even if the sensitive attribute is extracted from the feature vector, other features continue to affect predictions as a proxy of the sensitive attribute. One striking example was Amazon's recruiting AI case in 2018. Though Amazon erased gender from resumés before inserting them into the prediction model, the model downgraded proxies, such as careers including "women's" and graduation from two women's colleges.[188] Nevertheless, the use of ML, by itself, does not justify shifts in standards from disparate treatment to disparate impact. Moreover, U.S. courts do not construe discriminatory intent through counterfactual inferences and circumstantial evidence.[189] However, further examination is needed to determine whether a growing awareness of the proxy problem and the prevalence of tools for measuring disparate impact in ML models requires reconsideration of current standards.

Fourth, how would a system measure disparity under the output-centered approach? The notion of equality of opportunity is widely supported. Equality of outcome can be pursued through affirmative action when excessive inequalities exist. Equality of opportunity expects individuals with the same qualifications to have an equal chance of passing or failing, regardless of sensitive attributes including gender and race.

### (2) Tuning the Scope of Model Explanation

ADM laws are the primary statutory source requiring allocative or punitive AI developers to explain their models. However, the global proliferation of ADM laws raises concerns. ADM laws assume that people require additional safeguards, such as the ability to veto or receive explanations, when they are subjected to algorithms, as opposed to human decisions. This could result from a widespread conflation between automaticity and autonomy. While ADM laws are applicable when a decision is based solely on automation, the existence of such a system, aside from fully autonomous AI, is highly questionable, given that humans typically establish a sequence of preset operations or instructions.[190] It lacks substantiation that such automation leads to alienation or undermines human dignity or autonomy. Moreover, whether people have algorithmic aversion or algorithmic appreciation requires more substantiation. While experiments reveal people's aversion to the use of algorithm for forecasting student performance,[191] others show appreciation for the use of algorithm for numeric estimates regarding visual stimuli and for forecasts regarding the popularity of songs and romantic attraction.[192] The decision to provide individuals with the right to explanation should depend more on whether they are being categorized or evaluated for crucial tasks that significantly impact their lives over the long term, rather than simply whether an algorithm is being employed. It is also worth noting that human-AI collaboration can enhance transparency compared to solely manual processes.

---

[186] Washington v. Davis, 426 U.S. 229 (1976) (evaluating the Fourteenth Amendment). See also Justice Scalia's concurring opinion in Ricci v. DeStefano, 557 U.S. 557 (2009).
[187] Gillis & Spiess, *supra* note 1688, at 468–71.
[188] Jeffrey Dastin, *Amazon Scraps Secret AI Recruiting Tool that Showed Bias Against Women*, REUTERS, Oct. 11, 2018.
[189] Michael Selmi, *Proving Intentional Discrimination: The Reality of Supreme Court Rhetoric*, 86 GEO. L.J. 279, 291–94 (1997).

Furthermore, the notion of "algorithmic transparency" is flawed for ML, unlike in expert systems built on explicitly programmed algorithms. In ML, a model—not an algorithm—generally predicts, classifies, or generates something. Algorithms merely "fit" a model to a dataset pattern. Thus, the model's opaqueness originates from the underlying data distribution. In fact, most ML algorithms are transparent. And most source codes for ML, which embody an algorithm, are publicly available through libraries, including scikit-learn, TensorFlow, and PyTorch, or on public repositories like GitHub. When a hiring AI rejects an individual seeking a job, that person wants to know how each of their features influenced the outcome, not some quip that the "random forest algorithm was used to train and test the model."

Legislators often attempt to enact algorithmic transparency laws. However, the laws are not only misguided, but often contradict digital trade agreements, which prohibit parties from requiring the transfer of, or access to, source code or an algorithm expressed in that source code.[193]

---

[190] *See* Aziz Z. Huq, *A Right to a Human Decision*, 106 VA. L. REV. 611, 646 (2020); s*ee also* OFFICE OF MANAGEMENT AND BUDGET, M-24-10, MEMORANDUM FOR THE HEADS OF EXECUTIVE DEPARTMENTS AND AGENCIES (Mar. 28, 2024) at 29, https://www.whitehouse.gov/wp-content/uploads/2024/03/M-24-10-Advancing-Governance-Innovation-and-Risk-Management-for-Agency-Use-of-Artificial-Intelligence.pdf (noting that federal agencies must address "risks from the use of AI" regardless of whether a decision was fully automated, partially automated, merely informed by AI, or controlled or involved by humans).
[191] Berkeley J. Dietvorst, Joseph P. Simmons & Cade Massey, *Algorithm Aversion: People Erroneously Avoid Algorithms After Seeing Them Err*, 144 J. EXP. PSYCOL. GEN. 114 *passim* (2015).
[192] Jennifer M. Logg, Julia A. Minson & Don A. Moore, *Algorithm Appreciation: People Prefer Algorithmic to Human Judgment*, 151 ORG. BEHAV. HUM. DECIS. PROC. 90 *passim* (2019).

Transparency should instead refer to model explainability, meaning the degree to which a model's behavior can be explained in human terms. Predefining the context-dependent subject of explanation or disclosure poses a serious challenge for the law. A vague phrase—like the "logic involved in those decisionmaking processes" under the CPRA[194] or a requirement for AI systems to be "sufficiently transparent to enable deployers to interpret the system's output" under the E.U. AI Act[195]—might be the most detailed clarification possible. When implementing such vague phrases, it requires careful examination whether to require ex post on-demand explanation (typically based local explanation methods) or ex ante mandated disclosure (typically based on global explanation methods).

Another key factor to consider is gaming.[196] Once an AI model's parameters are disclosed, applicants are incentivized to manipulate their attributes to be favorably classified or scored. The transparency rule incentivizes applicants to game mutable features, impairing the classification models' accuracy. Gaming might also undermine fairness, because it can reduce the weight of mutable features and increases the weight of immutable features, including race, gender, and disability. Additionally, implementing mandatory disclosure of parameters can undermine privacy protections by empowering adversaries to single out or identify an individual's sensitive attributes through comparison of parameters fitted for a group to which the individual belongs and parameters fitted for a group from which the individual has departed.[197] Mandatory disclosure can further weaken security and safety by making the model more vulnerable to model stealing[198] or adversarial attacks.[199]

---

[193] *See, e.g.,* United States-Mexico-Canada Agreement, Dec. 10, 2019, art. 19 § 16; U.S.-Japan Digital Trade Agreement, Oct. 7, 2019, art. 17 § 1.
[194] CAL. CIV. CODE § 1798.185(a)(16) (2020).
[195] PE-CONS 24/24, art. 13(1).
[196] Jane Bambauer & Tal Zarsky, *The Algorithm Game*, 94 NOTRE DAME L. REV. 1 *passim* (2018); *see also* Ignacio N. Cofone & Katherine J. Strandburg, *Strategic Games and Algorithmic Secrecy*, 64 MCGILL L. J. 623 *passim* (2019) (arguing that concerns about gaming are over-blown,).
[197] CYNTHIA DWORK & AARON ROTH, THE ALGORITHMIC FOUNDATION OF DIFFERENTIAL PRIVACY 12–21 (2014).
[198] For the meaning of "model stealing" see Florian Tramèr et al., *Stealing Machine Learning Models via Prediction APIs*, USENIX SECUR. SYMP. (SEC) 601, *passim* (2016).
[199] For the meaning of "adversarial attacks" see Ian J. Goodfellow, Jonathon Shlens & Christian Szegedy, *Explaining and Harnessing Adversarial Examples*, ARXIV:1412.6572 *passim* (Mar. 20, 2015), https://arxiv.org/pdf/1412.6572.

Considering the above, the demands and expectations of those needing explanation should be carefully considered.

### D. Punitive AI

#### (1) Addressing Inaccuracy

Unlike allocative AI, punitive AI typically conducts absolute evaluation; thus, its major concern is the absolute accuracy rate, rather than the parity of accuracy rate across groups. For example, even if the rate of falsely convicted males is higher than that of females, the policy objective should target increased accuracy rates for both groups, rather than parity across groups. Disparity would therefore fall under representational harms to be examined from a long-term viewpoint, provided that an absolutely low accuracy rate causing disparity needs immediate intervention. Additionally, punitive AI must prioritize minimizing false positives—that is, wrongly convicted innocent people.

#### (2) Model Explanation

Punitive AI discussions can be approached in the same manner as allocative AI discussions.

### E. Cognitive AI

#### (1) Addressing QoS Deficiency and Disparity

One key issue for safety- or efficacy-related cognitive AI is ensuring safety and efficacy by guaranteeing a minimum level of QoS. QoS encompasses accuracy and robustness, as cognitive AI frequently operates in intended environments and unintended, but foreseeable, environments. A model is "robust" if it can maintain accuracy not only in source domains (i.e. where the training, test, and validation datasets were sourced) but also in different target domains (for example, in different natural conditions or under adversarial attacks).[200]

For self-driving vehicle sensors, robustness has emphasized ensuring adequate vehicle operation—even in situations distinct from the source domain. Techniques to improve robustness include: (1) multi-task learning, which "leverages useful information contained in multiple learning tasks to help learn a more accurate learner for each task"[201]; (2) domain adaptation, which adapts models trained from a source domain for a different target domain having a different distribution with its unlabeled data,[202] including fine-tuning pretrained language models; and (3) domain expansion, which adapts models for target domain, including domain adaptation, while maintaining the model's performance on the source domain.[203] This concept is yet to be formulated; guidelines must be proposed using industry standards.

---

[200] Eneko Agirre & Oier Lopez de Lacalle, *On Robustness and Domain Adaptation using SVD for Word Sense Disambiguation*, 22 COMPUT. LINGUIST. (COLING) 17, 17 (2008).

Recently, the normative focus of robustness has shifted to address QoS disparity arising from underrepresented minority groups in the dataset,[204] that is, the under-specification of subgroups.[205] As noted, self-driving cars are biased toward identifying pedestrians with light skin tones and against those with darker skin tones.[206] This issue has sparked discussions in medical imaging and diagnostics[207] where AI-enabled imaging and diagnostics attempt to score or classify patients' health conditions. As with punitive AI, cognitive AI's trustworthiness lies in absolute, not relative, accuracy and robustness. When data represents inaccurate demographics in clinical trials, diagnostic tools trained on such data make inaccurate predictions for underrepresented groups.[208] However, such QoS disparity problems differ from traditional discrimination issues, as seen in allocative AI.

---

[201] Yusuke Shinohara, *Adversarial Multi-task Learning of Deep Neural Networks for Robust Speech Recognition*, INTERSPEECH (2016).

[202] Ben-David et al., *supra* note 120, at 154.

[203] Jie Wang et al., *Unsupervised Domain Expansion for Visual Categorization*, 17 ACM T. MULTIM. COMPUT. COMMUN. APPL. 1 *passim* (2021).

[204] Blodgett et al., *supra* note 97, at 5456.

[205] Alexander D'Amour et al., *Underspecification Presents Challenges for Credibility in Modern Machine Learning*, 23 J. MACH. LEARN. RES. 1 *passim* (2022).

[206] *See generally,* Wilson et al., *supra* note 167.

[207] When categorizing medical AI as high-risk or critical, legislators might have postulated invasive treatment systems like AI-powered surgical robots. Despite great potential for it, the current use of AI in healthcare centers on non-invasive areas such as imaging and diagnostics. The top three medical specialties of 521 AI/ML-based SaMD approved by the U.S. Food and Drug Administration (FDA) until July 2022 are radiology (75%), cardiovascular (11%), and hematology (3%). FDA, *Artificial Intelligence and Machine Learning (AI/ML)-Enabled Medical Devices* (Oct. 5, 2022), https://www.fda.gov/medical-devices/software-medical-device-samd/artificial-intelligence-and-machine-learning-aiml-enabled-medical-devices. AI is being used for a wider range of medical purposes, including drug discovery and development, virtual assistants, and hospital management. In particular, AI designed to allocate access to medical service (such as hospital beds and emergency rooms) or medical products (such as vaccine and medicine) falls under allocative AI, but it does not account for a substantial portion of medical AI being used.

[208] Agostina J. Larrazabal et al., *Gender Imbalance in Medical Imaging Datasets Produces Biased Classifiers for Computer-Aided Diagnosis*, 117 PROC. NATL. ACAD. SCI. (PNAS) 12592 *passim* (2020).

Allocative AI should address the tradeoff in error rates of different groups, while attempting to equalize the error rates; cognitive AI should attempt to minimize the error rates of each group. While requiring allocative AI to be debiased prior to release is typically reasonable, such debiasing demands a more delicate balancing decision regarding whether to require cognitive AI to attain sufficient QoS for idiosyncratic patients, even at the cost of delaying its release to the public.

If cognitive AI is released before achieving sufficient QoS for minorities, minority communities must be duly warned. One relevant, albeit unrelated to AI, legal case involves Plavix (lopidogrel), an antiplatelet drug developed based on a clinical trial. In that legal case, ninety-five percent of participants were Caucasian, and the drug inadequately treated Pacific Islanders due to their greater genetic variation. [209] In March 2023, the Supreme Court of Hawaii held that the manufacturers violated Hawaii's Unfair or Deceptive Acts or Practices statute for failure to warn consumers about the poor responder issue.[210]

But there are limitations to including certain groups in clinical trials. South Korea's Ministry of Gender Equality and Family issued a guideline based on a 2017 gender impact assessment to increase female participation in phase 1 clinical trials. [211] However, female participation in clinical trials may be inherently insufficient because it should be avoided during childbearing or lactating years. Faced with data availability restrictions, developers should improve robustness—AI's ability to adequately perform in different domains—through robustness stress tests,[212] domain adaptation,[213] and data augmentation.[214]

In sum, addressing QoS problems and disparity in cognitive AI is not solely about eliminating biases, but also about ensuring AI performs optimally across different populations and contexts in terms of safety and efficacy.

### (2) Addressing Privacy Concerns

One key privacy issue is curbing AI-powered mass surveillance and maintaining bottom-up governance, allowing citizens to participate in reviewing misuse of their data. The E.U. AI Act bans AI use for (1) real-time remote biometric identification in public spaces for law enforcement, (2) emotion inference in the workplace or school, and (3) biometric categorization for deducing or inferring sensitive data.[215] From 2019 to 2021, Vermont, Virginia, and several U.S. municipalities passed legislation curbing facial recognition use.[216] However, amid a crime surge, many have undone or are undoing this restriction. Surveillance risks, which will be exacerbated by the transition to "smart cities," must be addressed through creative solutions, including privacy-enhancing technology and public-private partnerships.

---

[209] State ex rel. Shikada v. Bristol Myers Squibb Co., SCAP-21-0000363 (Haw. Mar. 15, 2023).
[210] *Id.*
[211] Ministry of Gender Equality and Family, *Gender Balance in Clinical Trials to be Attained for Safe Use of Medicine* (Aug. 18, 2017) (S. Kor.), https://www.mogef.go.kr/nw/enw/nw_enw_s001d.do?mid=mda700&bbtSn=705333.
[212] D'Amour et al., *supra* note 2055 *passim*.
[213] Ben-David et al., *supra* note 120, at 154.
[214] D'Amour et al., *supra* note 205 *passim*.

### (3) Other issues

For medical AI, transparency obligations should address how ethical duties require medical professionals to explain the basis of diagnoses and to obtain informed consent from patients. Generally, a doctor explaining something to a patient is best served by feature attributions in AI-based diagnostic imaging (e.g., identifying areas suggesting cancer with the highest probabilities).

For AI-based recommender systems, concerns regarding the ideological polarization caused by selective exposure or "filter bubble" have grown.[217] However, a solution to this representational harm requires a broader perspective based on further research regarding AI's impact on democracy.

## F. Generative AI
### (1) Coping with Toxic or Infringing Content Generation

Generative AI (genAI) can produce toxic languages or artwork, including infringing or otherwise unlawful content (e.g., obscene, defamatory, exploitive, copyright infringing, privacy infringing, suicide abetting, and fraudulent content). It can also be politically polarizing, present misleading market forecasts, or otherwise generate sensitive and provocative content.[218] GenAI can also be used to produce CBRN weapons, weapon blueprints, or malicious codes, undermining national security. Legislators must conduct careful risk assessments to balance social benefits, including those stemming from content moderation, and costs, including freedom of speech infringements.

---

[215] PE-CONS 24/24, art. 5(1)(f), (g), and (h).
[216] *See* 20 VT. STAT. ANN. § 4622 (2020); *see also* VA CODE § 23.1-815.1 (2022).
[217] *See generally* Marijn A. Keijzer & Michael Mäs, *The Complex Link between Filter Bubbles and Opinion Polarization*, 5 DATA SCIENCE 139 (2022).
[218] Hwaran Lee et al., *SQuARe: A Large-Scale Dataset of Sensitive Questions and Acceptable Responses Created Through Human-Machine Collaboration*, ASSOC. COMPUTE. LINGUISTICS (ACL) (2023).

Among the issues related to toxicity, the infringement of individual rights is the most problematic. GenAI's memorization and regurgitation of copyrighted materials and personal data raises serious concerns. Studies suggest that 0.007%[219] to 4.85%[220] of outputs from genAI are verbatim reproduction. The likelihood of a model regurgitating memorized data scales with model size, training data duplication, and prompt length.[221] Although several mitigation tools have been suggested—including de-duplication, data filtering, output filtering, instance attribution, differential privacy, and reinforcement learning from human feedback (RLHF)—issues persist, especially in copyright and privacy contexts.[222]

In the U.S., copyright holders have initiated many lawsuits against genAI developers.[223] Compared to text mining and the learning phase, the fair use doctrine or TDM exceptions are relatively less likely to apply to generative tasks. Some studies[224] are skeptical of the likelihood that courts will apply the fair use doctrine to genAI because genAI models can generate content resembling copyrighted data and their usage could replace economic markets beneficial to original creators.[225]

---

[219] Peter Henderson et al., *Foundation Models and Fair Use*, STAN. L. SCH. JOHN M. OLIN PROGRAM IN L. & ECON. WORKING PAPER SERIES 584, at 27 (2023) (citing Nikhil Kandpal, Eric Wallace & Colin Raffel, *Deduplicating Training Data Mitigates Privacy Risks in Language Models*, 162 INT. CONF. MAC. LEARN. (ICML) 10697 *passim* (2022)).

[220] *Id.* at 27. Jooyoung Lee et al., *Do Language Models Plagiarize? in* PROC. ACM WEB CONF. 2023 (2023).

[221] Nicholas Carlini et al., *Quantifying Memorization Across Neural Language Models*, ARXIV:2202.07646 *passim* (Mar. 6, 2023), https://arxiv.org/pdf/2202.07646; Kandpal et al., *supra* note 21819 *passim*.

[222] Henderson et al., *supra* note 21919, at 20–25.

[223] The plaintiffs include (1) book authors and guilds, *see e.g.* Tremblay v. OpenAI, Inc., No. 3:23-cv-03223 (N.D. Cal. filed Jun. 28, 2023); Kadrey et al. v. Meta Platforms, Inc., No. 3:23-cv-03417 (N.D. Cal. filed Jul. 7, 2023); Silverman et al. v. OpenAI, Inc., No. 3:23-cv-03416 (N.D. Cal. filed Jul 7, 2023); Chabon v. OpenAI, Inc., No. 3:23-cv-04625 (N.D. Cal. filed Sep 8, 2023); Authors Guild, et al. v. OpenAI, Inc. (S.D.N.Y. filed Sep. 19, 2023); Sancton v. OpenAI Inc., Microsoft Corporation, et al., No. 1:2023-cv-10211 (S.D.N.Y. filed Nov 21, 2023)); (2) visual artists, *see e.g.* Andersen et al. v. Stability AI Ltd., No. 23-cv-00201-WHO, 2023 WL 7132064 (N.D. Cal. Oct. 30, 2023); (3) an image database holder, Getty Images (US), Inc. v. Stability AI, Inc., No. 23-cv-00135 (D. Del. filed Feb. 3, 2023); (4) a recording company, *see* Concord Music Group, Inc. v. Anthropic PBC, No. 23-cv-01092 (M.D. Tenn. filed Oct. 18, 2023); (5) programmers, *see* Doe v. GitHub, Inc., No. 22-cv-06823 (N.D. Cal. filed Nov 3, 2022); Doe v. GitHub, Inc., No. 22-cv-07074 (N.D. Cal. filed Jan 26, 2023); and (6) various rightsholders, *see* J.L. et al., v. Google LLC et al., No. 23-cv-03440 (N.D. Cal. Jul. 11, 2023).

[224] Henderson et al., *supra* note 219, at 2 (citing Benjamin L.W. Sobel, *Artificial Intelligence's Fair Use Crisis*, 41 COLUM. J. L. ARTS 45 *passim* (2017); Mark A. Lemley & Bryan Casey, *Fair Learning*, 99 TEX. L. REV. 743, 777–78 (2020)).

[225] Henderson et al., *supra* note 219, at 2.

Moreover, TDM exceptions in each jurisdiction may not cover generation, particularly when they cover only TDM for non-commercial purposes or data mining only.

Accordingly, a new international framework should rebalance creation and access though alternatives, including (1) expanding the scope of fair use; (2) expanding and adjusting TDM exceptions to generative tasks; (3) introducing notice and takedown regimes, coupled with safe harbor regimes similar to the U.S. Online Copyright Infringement Liability Limitation Act,[226] which is applicable to online platforms; or (4) implementing a collective rights management (CRM), the collective collection of royalties by copyright societies, or private copying levy (PCL), a sales tax charged on purchases of media. CRM is akin to the collective management organizations (CMOs)'s collective collection of music royalties. A PCL solution could either resemble the PCL over digital audio recording devices and media under the U.S. Audio Home Recording Act of 1992,[227] or the PCL over music CD recorders under the U.S. Fairness in Music Licensing Act of 1998.[228] A jurisdiction's choice of alternative is likely to reflect the economic landscape and the interplay of conflicting interests between rightsholders and AI developers.

### (2) Deterring the Misuse of AI-Generated Synthetic Media

There are growing concerns regarding harms from the misuse of "deepfake" and other synthetic media generated by genAI for malicious purposes, including sex offenses, blackmailing, fraud, spear phishing, deceptive marketing, and misrepresenting well-known politicians during public elections. These misuses deceive consumers, investors, voters, the public, or reviewers. There are several types of deceptive misuses: (1) anthropomorphism, which involves pretending to be a fictitious person; (2) impersonation of a specific real person, including identity theft and ego extension; (3) counterfeiting others' work; and (4) AI-assisted plagiarism—all of which can be accompanied and amplified by disinformation.[229] Among these, impersonating another real person for fraud, marketing, or influencing public elections pose the most pressing concerns. Several U.S. states have already enacted legislation regarding the use of deepfakes in public elections.[230] Additionally, California's 2019 Bolstering Online Transparency (B.O.T.) Act requires the disclosure of non-human identity bots if such bots influence commercial transactions or election votes.[231]

---

[226] 17 U.S.C. § 512 (1998).
[227] 17 U.S.C. § 1008 (1992).
[228] 17 U.S.C. §§ 101, 110, 504 (1998).
[229] *See* Sungkyoung Jang, Eun Seo Jo & Sangchul Park, *The Misuse of Generative AI for Deception: Status Quo, Mitigating Technologies, and Legal Challenges*, 22 KOR. ECON. L. REV. 3, 8–16 (2023) (S. Kor.).

Lawmakers should focus on deterring deceptive misuses and implementing penalties for attempts to circumvent technological safeguards. Further, to mitigate social costs and provide recipients with accessible tools to evade deception, genAI providers should be expected to adopt technological measures, including watermarking; digital platforms offer corresponding detection tools.[232]

Several jurisdictions have already imposed obligations to take technical measures. In the United States, E.O. 14110 requires the DOC to prepare a report on (1) provenance and authentication; (2) labeling, including watermarking; (3) detecting synthetic content; (4) preventing child sexual abuse or deepfakes; (5) testing software used for the purposes of (1) to (4); and (6) auditing and maintaining synthetic content, and develop guidance regarding existing tools and practices.[233] The Office of Management and Budget is further tasked with preparing guidance to agencies for labeling and authenticating digital content that they produce or publish.[234]

China requires genAI providers to add watermarking to synthetic media and use labeling if the media is likely to arouse confusion or misunderstanding.[235]

The E.U.'s Digital Services Act obliges providers of "very large online platforms" and search engines to ensure synthetic media is distinguishable through prominent markings when presented online and to provide easy functionality for recipients indicating such information.[236] The AI Act stipulates that genAI providers should ensure the outputs of the AI system are marked in a machine-readable format and detectable as artificially generated or manipulated.[237]

---

[230] *See, e.g.,* TEX. ELEC. CODE ANN. § 255.004(d) (2021); CAL. ELEC. CODE § 20010(a) (2020, amended 2022); MINN. STAT. §§ 609.771, 617.262 (2023); and WASH. REV. CODE. § 42.62 (2023).
[231] CAL. BUS. & PROF. CODE § 17941 (2018).
[232] Jang, Jo & Park, *supra* note 229, at 16–20.
[233] § 4.5(a) and (b). *See also* NIST, *supra* note 62.
[234] *Id.* § 4.5(c).
[235] Hulianwang Xinxi Fuwu Shendu Hecheng Guanli Guiding [Internet Information Service Deep Synthesis Management Provisions] (promulgated by the Cyberspace Admin. China, the Min. Ind. & Info. Tech., and the Min. Publ. Sec., Nov. 25, 2022, effective Jan. 10, 2023), arts. 16 and 17.
[236] *See* art. 35(1)(k).

Provenance and authentication tools, particularly text watermarking, are nascent technologies that require improvement. Thus, facilitating self-regulation or incentivizing implementation through mechanisms, including safe harbor provisions, are more suitable than imposing strict legal obligations.[238] Additionally, it might be reasonable for genAI providers to promptly take measures upon receiving notices of deceptive actions, including shutting down or limiting APIs, and providing relevant information. Clear procedures should also be implemented to handle affected parties' requests to halt the processing of personal data, including their likeness.[239]

### (3) Other Issues

Though less imminent than infringement and misuse, newly arising issues involving genAI must be reviewed from a long-term perspective. "Hallucination," disclosure of non-human identity, and stereotyping rank among the more pressing concerns.

Language models, particularly auto-regressive language models like GPT, generate each token that is most likely to follow preceding tokens. During that process, they are often observed to generate factually incorrect but seemingly plausible or confidently stated statements, a phenomenon called "hallucination"[240] or "confabulation."[241] To combat hallucination, developers have created tools combining the power of language models with external dynamic knowledge bases, including service engines[242] and online encyclopedias.[243] While hallucination does not merit immediate legal intervention, there may be a stronger argument for requiring disclosure of the factual basis of the information where there is a potential for harm or significant consequences based on the information being generated. This is especially the case with regard to medical, legal, or other professional advice or asset price forecasts, for example, by robo-advisers.

---

[237] PE-CONS 24/24, art. 50(2).
[238] Jang, Jo & Park, *supra* note 229, at 26–29.
[239] *Id.* at 29–30.
[240] *See generally* Nouha Dziri et al., *On the Origin of Hallucinations in Conversational Models: Is it the Datasets or the Models?*, ASSOC. COMPUT. LINGUISTICS (ACL) 5271 (2022).
[241] NIST, NIST AI 600-1, ARTIFICIAL INTELLIGENCE RISK MANAGEMENT FRAMEWORK: GENERATIVE ARTIFICIAL INTELLIGENCE PROFILE, Initial Public Draft at 5 (2024).
[242] *See generally* Jacob Menick et al., *Teaching Language Models to Support Answers with Verified Quotes*, ARXIV:2203.11147 (Mar. 21, 2022), https://arxiv.org/abs/2203.11147.
[243] Reiichiro Nakano et al., *WebGPT: Browser-Assisted Question-Answering with Human Feedback*, ARXIV:2112.09332 *passim* (Jun. 1, 2022), https://arxiv.org/pdf/2112.09332.

There have also been discussions about whether individuals interacting with chatbots or other genAI should be informed that their interaction is with a non-human, even in cases not involving deceptive misuse. Under the E.U. AI Act, providers must design and develop AI systems interacting with people such that individuals are informed they are interacting with an AI.[244]

However, the benefits of disclosing the non-human identity are questionable. Empirical studies on the impact of AI-mediated communication (AI-MC) on consumer trust [245] make the following observations: (1) chatbot disclosure negatively influences consumer trust;[246] (2) the more human a service is *perceived* to be, the greater humans trust it, regardless of whether that service is a chatbot or a human customer services representative;[247] and (3) Airbnb customers place greater trust in AI-written host profiles compared to a mixed set of AI- and human-written host profiles. [248] These studies suggest that, unless necessary for preventing deceptive misuse, mandatory non-human identity disclosures require clearer policy justifications rather than mere conjectures regarding human autonomy.

Stereotyping in generation models can strengthen and reproduce over-generalized beliefs about a particular group. Several tools have been proposed to measure and address stereotypes: some studies compute differences in a word's cosine similarities (e.g., he/she or man/woman) to attribute bias[249]; others use stereotype scores, the "percentage of examples in which a model prefers a stereotypical association over an anti-stereotypical association."[250] To debias language models, some propose counterfactual data augmentation, which swaps bias attributes in a dataset and retrains a model.[251] Others conduct sequential iterative nullspace projections to obtain a more uniform distribution over bias attributes.[252] The literature tends to assume that large language models (LLMs) must be completely devoid of biases. However, if the corpus used to train the LLM is sufficiently representative of the population, stereotyping occurs not due to biases themselves, but rather from the decoder's behavior in showing only the token based on the highest probability. If the decoder randomly generates tokens proportionate to each token's probability of occurrence, stereotyping or underrepresentation might not occur. One exception is where outdated, historical datasets that do not reflect recent social changes are fed into an LLM. There, the LLM might not duly represent the population. These situations merit active debiasing effort. The intricate and contextual nature of stereotyping suggests that a long-term review—rather than immediate legal intervention—is more appropriate to help developers hone measuring and debiasing tools and facilitate self-regulation. However, if stereotypes transition to allocative harms in downstream tasks due to fine-tuning,[253] discussions regarding discriminative models may be applicable.

---

[244] PE-CONS 24/24, art. 50(1).
[245] Jeffrey T. Hancock, Mor Naaman & Karen Levy, *AI-Mediated Communication: Definition, Research Agenda, and Ethical Considerations*, 25 J. COMPUT.-MEDIAT. COMMUN. 89, 96 (2020).
[246] *See generally* Nika Mozafari, Welf H. Weiger & Maik Hammerschmidt, *The Chatbot Disclosure Dilemma: Desirable and Undesirable Effects of Disclosing the Non-Human Identity of Chatbots*, 41 INT. CONF. INF. SYST. (ICIS) 1 (2020).
[247] *See generally* Lincoln Lu et al., *Measuring Consumer-Perceived Humanness of Online Organizational Agents*, 128 COMPUT. HUM. BEHAV. 107092 (2022).
[248] *See generally* Maurice Jakesch et al., *AI-Mediated Communication: How the Perception that Profile Text was Written by AI Affects Trustworthiness*, CONF. HUM. FACTORS COMPUT. SYST. (CHI) 1 (2019).
[249] *See generally* Aylin Caliskan, Joanna J. Bryson & Arvind Narayanan, *Semantics Derived Automatically from Language Corpora Contain Human-like Biases*, 356 SCIENCE 183 (2017); *see also* Chandler May et al., *On Measuring Social Biases in Sentence Encoders*, ASSOC. COMPUT. LINGUISTICS (ACL) 622 (2019).

## 4. Commensurability: Formulating Metrics Using Industry Standards

### A. Overview

Transitioning to a principle-based approach is an ineffective alternative to dense and inflexible horizontal rules like the E.U. AI Act. A valid alternative should reduce stringency and lack of granularity, rather than foreseeability and commensurability. To achieve this, coherent policy objectives addressing the tangible harms of AI types, as established in paragraph 3, should be transformed into quantifiable, measurable, and machine-readable metrics using international industry standards. As noted, this can be attained by decentralized governance more efficiently.

---

[250] Moin Nadeem, Anna Bethke & Siva Reddy, *StereoSet: Measuring Stereotypical Bias in Pretrained Language Models*, ASSOC. COMPUT. LINGUISTICS (ACL) 5356, 5361 (2021).
[251] *See generally* Ran Zmigrod et al., *Counterfactual Data Augmentation for Mitigating Gender Stereotypes in Languages with Rich Morphology*, ASSOC. COMPUT. LINGUISTICS (ACL) 1651 (2019).
[252] *See generally* Paul P. Liang et al., *Towards Understanding and Mitigating Social Biases in Language Models*, INT. CONF. MACH. LEARN. (ICML) 6565 (2021).
[253] *See generally* Mohsen Abbasi et al., *Fairness in Representation: Quantifying Stereotyping as a Representational Harm*, SIAM INT. CONF. DATA. MIN. (SDM) 801 (2019).

This approach offers several advantages. First, it enables clear and unambiguous discussions based on well-defined mathematical criteria, avoiding linguistic confusion and ensuring shared understanding among stakeholders worldwide. Second, quantifiable, measurable metrics are readily integrated into AI systems during development, particularly after being materialized into an open-source library. Such "compliance by design" also effectively applies to deployers and users, resulting in cost-efficient, yet effective, compliance across the supply chain. Third, consensus on metrics based on industry standards will facilitate the defragmentation of local regulations over time. Fourth, scalability and flexibility are ensured because industry groups are empowered to tailor and adjust metrics to their specific needs and contexts.

Industry standards should encompass various fairness, accuracy, and robustness metrics, and other standards that shape policy objectives. Within two years of the E.U. AI Act's enactment, the European Committee for Standardization (CEN) and the European Committee for Electrotechnical Standardization (CENELEC) will collaborate to develop harmonized standards (HAS).[254] However, the scalability of this effort may be constrained by the E.U. AI Act's stringent horizontal framework. In such case, the NIST's AI RMF and ISO/IEC JTC 1/SC42 standards could serve as a potential starting point for the global pursuit of commensurability.

### B. Accuracy and Fairness Metrics

For regressors dealing with continuous outcomes, mean squared errors are commonly used as accuracy metrics. For binary or multinomial classifiers, accuracy refers to predictive accuracy, representing the proportion of correct predictions among all predictions. Predictive accuracy often assumes an equal weight on false positives and false negatives. However, this approach is sometimes inappropriate, especially when classifying individuals. In such cases, accuracy metrics must be decomposed by comparing observed results and the ground truth.

While equality in outcome may conflict with accuracy, equality in opportunities begins with accuracy. Specifically, achieving parity of these accuracy rates, or error rates, among groups becomes crucial for promoting equality in opportunities. Details are as follows.

### (1) Allocative AI

---

[254] PE-CONS 24/24, arts. 40 and 41.

Among accuracy metrics, allocative AI should prioritize *sensitivity*. Sensitivity is equivalent to the ratio of true positives to actual positives. For instance, it reflects the proportion of competent students who pass the exam. That aims to minimize false negatives, including competent but unsuccessful applicants. This assumption is based on the belief that unsuccessful applicants' undeserving failures outweigh a windfall from successful but incompetent applications. However, where applicants are assigned to important public posts impacting many people, false positives must also be minimized. In such case, *specificity*, which equals to the ratio of true negatives to actual negatives, is more meaningful.

Since allocative AI performs relative evaluations in most cases, its objective should be to achieve parity in accuracy rates. Parity in sensitivity, often referred to as *equal opportunity*,[255] would serve as a suitable fairness metric to gauge allocative fairness.

### (2) Punitive AI

Among accuracy metrics, punitive AI should emphasize *precision*. Precision is equivalent to the ratio of true positives (real criminals) to the total number predicted positive (convicts). This emphasis on precision aligns with the fundamental requirement for criminal justice: the presumption of innocence, which aims to maximize precision (or minimize wrongfully convictions) at the expense of the proportion of individuals wrongfully acquitted. However, in exceptional cases, where extremely dangerous criminals should be detected and apprehended with urgency, *negative predictive value (NPV)* gains relatively more importance. NPV represents the ratio of true negatives (real innocents) to the total number predicted negative (individuals acquitted).

Since punitive AI performs absolute evaluations in most cases, its objective should be to achieve *a certain absolute level of accuracy rates in each group*. In other words, a sufficiently high level of precision is crucial for punitive AI.

Controversy has surrounded whether the Correctional Offender Management Profiling for Alternative Sanctions (COMPAS), a recidivism-prediction tool used in New York, Wisconsin, California, and other U.S. states for bail and sentencing decisions, is racially biased. COMPAS assigned defendants scores on a 1–10 scale, predicting their reoffence rate.[256] In 2016, ProPublica obtained data from Broward County, Florida, through a Freedom of Information request. The data included the recidivism score that COMPAS predicted and whether defendants had actually recidivated over two years.[257] According to ProPublica, "black defendants who did not recidivate over a two-year period were nearly twice as likely to be misclassified as higher risk compared to their white counterparts (45 percent vs. 23 percent)."[258] And "white defendants who re-offended within the next two years were mistakenly labeled low risk almost twice as often as black re-offenders (48 percent vs. 28 percent)."[259] In response, COMPAS developer Northpointe (currently, Equivalent) counterargued that COMPAS was well-calibrated because, among defendants who scored the same on the COMPAS scale, the same percentage of white and black defendants re-offended.[260]

---

[255] *See* Moritz Hardt, Eric Price & Nati Srebro, *Equality of Opportunity in Supervised Learning,* 29 ADVANCES NEURAL INF. PROCESS. SYST. 3323, 3324–25 (2016).

ProPublica and Northpointe based their arguments on different fairness metrics, respectively. Table 1 shows the distribution of black and white defendants in the Broward County data that ProPublica obtained. A defendant's recidivism risk is "high" if the COMPAS score is 4 or more. Otherwise, it is "low."

|  | All Defendants | | Black Defendants | | White Defendants | |
| --- | --- | --- | --- | --- | --- | --- |
|  | High | Low | High | Low | High | Low |
| Recidivated | 2,035 | 1,216 | 1,369 | 532 | 505 | 461 |
| Survived | 1,282 | 2,681 | 805 | 990 | 349 | 1,139 |

Table 1: Confusion matrix regarding the COMPAS score.[261]

In this example, ProPublica highlighted that sensitivity was higher in black defendants, while specificity was higher in white defendants:[262]

---

[256] Jeff Larson et al., *How We Analyzed the COMPAS Recidivism Algorithm*, PROPUBLICA (May 23, 2016), https://www.propublica.org/article/how-we-analyzed-the-compas-recidivism-algorithm.
[257] *Id.*
[258] *Id.*
[259] *Id.*
[260] William Dieterich, Christina Mendoza & Tim Brennan, COMPAS RISK SCALES: DEMONSTRATING ACCURACY EQUITY AND PREDICTIVE PARITY (2016), https://go.volarisgroup.com/rs/430-MBX-989/images/ProPublica_Commentary_Final_070616.pdf.
[261] Larson et al., *supra* note 256.
[262] *Id.*

> *Sensitivity* (the probability of reoffenders being classified as high risk):
>   1369 / (1369 + 532) = 0.72 or 72% for black defendants
>   505 / (505 + 461) = 0.52 or 52% for white defendants
> *Specificity* (the probability of non-reoffenders being classified as low risk):
>   990 / (805 + 990) = 0.55 or 55% for black defendants
>   1139 / (349 + 1139) = 0.77 or 77% for white defendants

In contrast, Northpointe argued precision and NPV were similar across races:[263]

> *Precision* (the probability of those classified as high-risk reoffending):
>   1369 / (1369 + 805) = 0.63 or 63% for black defendants
>   505 / (505 + 349) = 0.59 or 59% for white defendants
> *NPV* (the probability of those classified as low-risk not reoffending):
>   990 / (532 + 990) = 0.65 or 65% for black defendants
>   1139 / (461 + 1139) = 0.71 or 71% for white defendants

As this case illustrates, differences in the base rate of reoffence among groups create discrepancies when evaluating fairness metrics from the viewpoint of ground truth (the parity of sensitivity and specificity) and observation (the parity of precision and NPV). As noted, given that the focus of criminal justice has been on precision, the observation-based approach can make more sense.

In fact, regardless of which metric holds more merit, the use of fairness metrics, rather than accuracy metrics, is misplaced. The critical consideration should not be a relevant comparison but the absolute level of prediction. With a precision of only fifty-nine percent for white defendants and sixty-three percent for black defendants, the system's performance is unacceptably low to justify its use in criminal justice. Forty-one percent of the white defendants and thirty-seven percent of the black defendants were classified as high risk, though they did not reoffend within two years.

---

[263] Dieterich, Mendoza & Brennan, *supra* note 260, at 11.

### (3) Cognitive AI

Appropriately balancing sensitivity and specificity for cognitive AI depends on the specific safety and efficacy concerns involved. For instance, for a self-driving car sensor's recognition of pedestrians, high sensitivity is required on average and across diverse demographics, including those based on skin color, age group, and disability. Stress testing is therefore needed to ensure robustness. Similarly, sensitivity is crucial for detecting cancer using medical imaging and diagnostics, as sensitivity directly affects the ability to detect potential cases. Specificity should also be carefully considered to mitigate unnecessary risks, as where false positives—in diagnostic procedures such as biopsies—could lead to risky follow-up procedures.

### C. Explanation Standards

Self-interpretable models are explained by disclosing parameters. For example, a regression model predicting weight, given height and waist size, can be explained as "predicted weight increases by one whenever height increases by $a$ or waist size increases by $b$." For less interpretable models like neural networks or ensemble models, explainable AI (XAI) techniques have been devised to improve *post-hoc* interpretability. Industries must devise reasonable standards for the scope and method of explanation based on XAI techniques. XAI is categorized into (1) global model-agnostic methods, describing expected values based on data distribution; (2) local model-agnostic methods, providing easier interpretations of individual predictions; and (3) model-specific methods explaining a certain scope of models.[264]

If a law requires ex post explanation based on a specific decision-making, local model-agnostic methods can become a candidate for an industry standard. The Shapley value, which produces the "average marginal contribution of a feature value across all possible coalitions," is regarded as one of the most consistent and solid methods, providing predictions that are fairly distributed along feature values.[265] However, robustness alone does not ensure a good explanation. Explainability should reflect the needs and cognition of those needing an explanation.[266] As such, some view instance-based explanations—particularly counterfactual explanations (CFE)—as an alternative solution.[267] CFEs describe the "smallest change to the feature values that changes the prediction to a predefined output."[268] For example, assume that John was disqualified for a loan application because he had too many credit cards. John might be curious about how many credit cards he needs to cancel to obtain a sufficient credit score. That is exactly what CFE does.

---

[264] CHRISTOPHER MOLNAR, INTERPRETABLE MACHINE LEARNING: A GUIDE FOR MAKING BLACK BOX MODELS EXPLAINABLE, 2nd ed., 109 – 296 (2022).

[265] *Id*. at 215–17, 224.

[266] David A. Broniatowski, *Psychological Foundations of Explainability and Interpretability in Artificial Intelligence,* NIST INTERAGENCY REPORT NO. 8367 (2021).

If a law requires ex ante explanation, global model-agnostic methods become a candidate for an industry standard. Global surrogate model, partial dependence plot (PDP),[269] accumulated local effects (ALE),[270] or maximum mean discrepancy (MMD)-critic[271] could be utilized to measure the impact of a certain feature on the prediction on average.

## V. CONCLUSION

Thus far, this paper has explored a strategy to address fragmentation potentially stemming from a global divide in AI regulations. Among various solutions, the paper critiques the horizontal framework, likening it to the risk posed by the Locomotive Act of 1865, which imposed a speed limit of 2 to 4 mph based on a zero-risk mindset and hindered the potential for novel technology to benefit society. With context-specific approaches as a starting point, the paper aims to propose a more coherent and commensurable framework to foster global harmonization and interoperability. Its purpose is not solely to alleviate undue regulation that stifles innovation, but also to precisely target and effectively mitigate potential harms. In this endeavor, the paper seeks to identify legal safeguards akin to safety belts, which not only save lives but also empower drivers to move with greater confidence and agility. Specifically, if a decentralized governance system is established, the proposed 3C framework is anticipated to align effectively with it. We can achieve this tricky mission only by approaching AI regulation with the same level of diligence and expertise that humanity currently invests in AI development.

---

[267] *See generally* Sandra Wachter, Brent Mittelstadt & Chris Russell, *Counterfactual Explanations Without Opening the Black Box: Automated Decisions and the GDPR*, 31 HARV. J. L. TECH. 841 (2018).
[268] Molnar, *supra* note 264, at 194.
[269] *See generally* Jerome H. Friedman, *Greedy Function Approximation: A Gradient Boosting Machine*, 29 ANN. STAT. 1189 (2001).
[270] *See generally* Daniel W. Apley & Jingyu Zhu, *Visualizing the Effects of Predictor Variables in Black Box Supervised Learning Models*, 82 J. R. STAT. SOC. SER. B STAT. METHODOLOGY 1059 (2020).
[271] *See generally* Been Kim, Rajiv Khanna & Sanmi Koyejo, *Examples are not Enough, Learn to Criticize! Criticism for Interpretability*, 29 ADVANCES IN NEURAL INF. PROCESS. SYST. 2288, 2288–96 (2016).